\pdfoutput=1

\documentclass[11pt]{article}

\usepackage[final]{acl}

\usepackage{times}
\usepackage{latexsym}

\usepackage[T1]{fontenc}

\usepackage[utf8]{inputenc}

\usepackage{microtype}

\usepackage{inconsolata}

\usepackage{graphicx}

\usepackage{hyperref}
\usepackage{url}
\usepackage{amsfonts}
\usepackage[utf8]{inputenc} 
\usepackage[T1]{fontenc}    
\usepackage{booktabs}
\usepackage{graphicx}
\usepackage{multirow}
\usepackage{multicol}
\usepackage{enumitem}
\usepackage{microtype}
\usepackage{caption}
\usepackage{subcaption}
\usepackage{wrapfig}
\usepackage{listings}
\usepackage{csquotes} 
\usepackage{color}
\usepackage{xcolor} 
\usepackage{colortbl} 
\usepackage{framed} 
\usepackage{amsmath}
\usepackage{amssymb}
\usepackage{mathtools}
\usepackage{amsthm}
\usepackage{arydshln} 
\usepackage{algorithm}
\usepackage{algorithmic}
\usepackage[capitalize,noabbrev]{cleveref}

\usepackage{etoolbox}
\AtBeginEnvironment{multline}{
\setlength{\abovedisplayskip}{2pt}
\setlength{\belowdisplayskip}{2pt}
\setlength{\abovedisplayshortskip}{1pt}
\setlength{\belowdisplayshortskip}{1pt}
}
\usepackage[utf8]{inputenc}
\usepackage{color}
\usepackage{colortbl}

\usepackage{tabularx, array}
\newcolumntype{C}[1]{>{\centering\arraybackslash}p{#1}}
\usepackage{hyphenat}
\newcolumntype{Y}{>{\centering\arraybackslash}X} 
\usepackage[most]{tcolorbox}
\tcbuselibrary{breakable} 
\usepackage{fancyvrb}    
\tcbuselibrary{breakable,skins}   
\tcbuselibrary{listings, breakable}
\usepackage{enumitem}  
\usepackage{fvextra}   
\usepackage{booktabs}
\usepackage{pifont}
\usepackage{multirow}
\DefineVerbatimEnvironment{Verbatim}{Verbatim}{breaklines=true, breakanywhere=true}
\definecolor{ysdarkpurple}{HTML}{4E2399}
\definecolor{ysshallowpurple}{HTML}{E6DBFF}
\definecolor{ysdarkred}{HTML}{8c2824}
\definecolor{ysshallowred}{HTML}{F8D7D7}
\definecolor{ysdarkblue}{HTML}{005E99}
\definecolor{ysshallowblue}{HTML}{CCEBFF}
\definecolor{ysdarkgrey}{HTML}{333333}
\definecolor{ysshallowgrey}{HTML}{E5E5E5}
\newtcbinputlisting{\promptbox}[2][]{
  enhanced,
  breakable, 
  colback=ysshallowblue,
  colframe=ysdarkblue,
  fonttitle=\bfseries,
  title=#2,
  listing only, 
  listing options={
    language=YAML,          
    basicstyle=\ttfamily\small, 
    breaklines=true,        
    breakatwhitespace=true, 
    postbreak=\mbox{\textcolor{red}{$\hookrightarrow$}\space}, 
    showstringspaces=false, 
    upquote=true,           
  },
  #1 
}
\definecolor{ColorGrok}{HTML}{FFFDE7}      
\definecolor{ColorPplx}{HTML}{EFFDFE}      
\definecolor{ColorOpenAI}{HTML}{F2F2F2}    
\definecolor{ColorGemini}{HTML}{E6F4FE}    
\definecolor{ColorClaude}{HTML}{FFF3EB}    
\definecolor{SectionHeaderColor}{HTML}{FFFFFF} 
\colorlet{DarkerColorClaude}{ColorClaude!95!black}
\colorlet{DarkerColorPplx}{ColorPplx!95!black}
\colorlet{DarkerColorGemini}{ColorGemini!95!black}
\colorlet{DarkerColorOpenAI}{ColorOpenAI!95!black}
\colorlet{DarkerColorGrok}{ColorGrok!95!black}

\newcommand{\ours}{Meta-Reasoner}

\newcommand{\refsec}[1]{\S\ref{#1}} 

\title{Meta-Reasoner: Dynamic Guidance for Optimized Inference-time \\Reasoning in Large Language Models}

\makeatletter
\renewcommand*{\@fnsymbol}[1]{\ensuremath{\ifcase#1\or *\or \dagger\or \ddagger\or
   \mathsection\or \mathparagraph\or \|\or **\or \dagger\dagger
   \or \ddagger\ddagger \else\@ctrerr\fi}}
\makeatother

\author{
    \textbf{Yuan Sui}\textsuperscript{1}\thanks{\ Corresponding Email: \texttt{yuan.sui@u.nus.edu}} \quad
    \textbf{Yufei He}\textsuperscript{1} \quad
    \textbf{Tri Cao}\textsuperscript{1} \quad
    \textbf{Simeng Han}\textsuperscript{2} \quad
    \textbf{Yulin Chen}\textsuperscript{1} \quad
    \textbf{Bryan Hooi}\textsuperscript{1} \\
    \textsuperscript{1}National University of Singapore \quad
    \textsuperscript{2}Yale University
}

\begin{document}
\maketitle

\begin{abstract}

Large Language Models (LLMs) often struggle with computational efficiency and error propagation in multi-step reasoning tasks. While recent advancements on prompting and post-training have enabled LLMs to perform step-wise reasoning, they still tend to explore unproductive solution paths without effective backtracking or strategy adjustment. In this paper, we propose Meta-Reasoner, a new framework that empowers LLMs to ``think about how to think''. It optimizes the inference process by dynamically adapting reasoning strategies in real-time. Our approach employs contextual multi-armed bandits (CMABs) to learn an adaptive policy. It learns to evaluate the current state of LLM's reasoning and determine optimal strategy that is most likely to lead to a successful outcome during inference, like whether to backtrack, switch to a new approach, or restart the problem-solving process. This meta-guidance helps avoid unproductive paths exploration during inference and hence improves computational efficiency. 
We evaluate Meta-Reasoner on math problems (e.g., Game-of-24, TheoremQA) and scientific tasks (e.g., SciBench). Results show that our method outperform previous SOTA methods by 9-12\% in accuracy, while reducing inference time by 28-35\% under the same compute budget. Additional experiments on creative writing demonstrate the generalizability of our approach to diverse reasoning-intensive tasks.
\footnote{Code are released at \url{https://github.com/Y-Sui/Meta-reasoner}}

\end{abstract}
\section{Introduction}

Recent advancements on post-training of large language models (LLMs) like o1/o3/r1~\cite{openai2024openai,deepseek-ai2025deepseekr1}, have achieve remarkable performance on complex reasoning tasks, such as math~\cite{patel2024aimeaisystem,lightman2023letsverifystep}, science~\cite{rein2023gpqagraduatelevelgoogleproof}, and logical puzzles~\cite{lei2024macmutilizingmultiagent,yao2023treethoughtsdeliberate}.
By simulating human-like deliberation~\cite{yao2024mulberry0,wei2022chain}, these approaches enable LLMs to decompose problems into subproblems, test hypotheses, reflect on intermediate results, and iteratively refine candidate solutions at inference time~\cite{cao2024surveylargelanguage}.
This extended inference-time reasoning allows models to progressively improve intermediate solutions before committing to a final answer~\cite{chenghaoyang2024inferencetimecompute}.

Despite these advances, this deliberate inference-time reasoning remains fundamentally trial-and-error. While this facilitates exploration of diverse solution strategies, recent works on scaling test-time compute show that naively increasing the amount of reasoning (e.g., the number of generated tokens or steps) leads to diminishing returns, where accuracy plateaus and stops improving despite the extra computational cost~\cite{snell2024scalingllmtesttime,manvi2024adaptiveinferencetimecompute}.
Specifically, current approaches face two key limitations: (1) \textbf{computational inefficiency}, where models expend substantial inference-time compute on unproductive or redundant reasoning trajectories~\cite{chenghaoyang2024inferencetimecompute}; and (2) \textbf{error propagation}, where early mistaken assumptions propagate through long chains of reasoning and are difficult to revoke~\cite{lei2024macmutilizingmultiagent,ling2023deductiveverificationchainofthought}. Empirical studies indicate that longer chain-of-thought (CoT) traces can improve accuracy, but they are increasingly expensive and prone to unproductive loops in the absence of higher-level guidance~\cite{havrilla2024glorewhenwhere}.

Recent techniques such as backtracking and self-reflection can partially mitigate these issues by revising or critiquing model outputs~\cite{gandhi2024streamsearchsos,li2025searcho1agenticsearchenhanced}. However, these methods still lack a principled mechanism to \emph{holistically} revise or redirect the ongoing reasoning process. In particular, they provide limited support for dynamically changing the overall strategy (e.g., choosing to abandon an unpromising decomposition or restart from alternative premises) when the current trajectory is unlikely to succeed~\cite{gao2024metareasoninglarge}. Additionally, as LLMs are tackling more challenging problems which require longer reasoning, this absence of adaptive, higher-level control becomes more critical. \textbf{A critical challenge}, therefore, is to enable LLMs to manage their reasoning budget more effectively, i.e., prioritizing promising directions while adapting or discarding ineffective strategies during inference time. Addressing this requires a novel approach to provide adaptive oversight of the reasoning process.

In this paper, we propose \ours{}, a meta-reasoning module that operates alongside an LLM to dynamically optimize its reasoning strategy during inference. Acting as a high-level controller, \ours{} continuously evaluates the LLM’s current reasoning state and provides strategic guidance. Inspired by dual-process theory~\cite{didolkar2024metacognitivecapabilitiesllms}, we explicitly decouple \emph{high-level strategy selection} (System-2-like) from \emph{low-level stepwise generation} (System-1-like) via a lightweight \emph{progress report} interface.
In specific, \ours{} only considers a compact summary of the LLM’s recent reasoning and then proposes updated strategies based on this summary. The system operates iteratively: (1) the base LLM generates partial CoT reasoning steps (\refsec{sec:cot_generation}) and summarizes its current reasoning state into a lightweight progress report (\refsec{sec:progress_report}). 
(2) \ours{} reviews this report and uses a contextual multi-armed bandit (CMAB) algorithm (\refsec{sec:strategy_generation}) to select high-level actions (e.g., backtrack to a previous step, change the decomposition scheme, or restart from alternative axioms);
(3) the LLM then continues reasoning, conditioned on both the original context and the controller’s guidance. This design allows \ours{} to focus on strategic control while leaving fine-grained token-level generation to the base model, thereby reducing overhead and improving robustness.
Based on our experiments results (\refsec{sec:experiments}), we find that \ours{} consistently improves performance over strong inference-time reasoning baselines on complex mathematical, logical, and puzzle benchmarks, and that its meta-reasoning strategies transfer to other domains such as creative writing.
\textbf{Overall, our main contributions are:}
\begin{itemize}[leftmargin=*]
\setlength\itemsep{0em}
    \item We introduce a meta-reasoning framework that provides high-level strategic control over LLM inference, reducing the tendency of models to get trapped in unproductive reasoning trajectories.
    \item We design a lightweight progress reporting mechanism that enables efficient communication of the LLM’s internal reasoning state to the meta-reasoner with minimal inference-time compute.
    \item We demonstrate that \ours{} requires no task-specific fine-tuning and generalizes across domains, including mathematics, science, puzzles, and open-ended creative problems.
\end{itemize}

\begin{figure}[t]
    \centering
    \includegraphics[width=\linewidth]{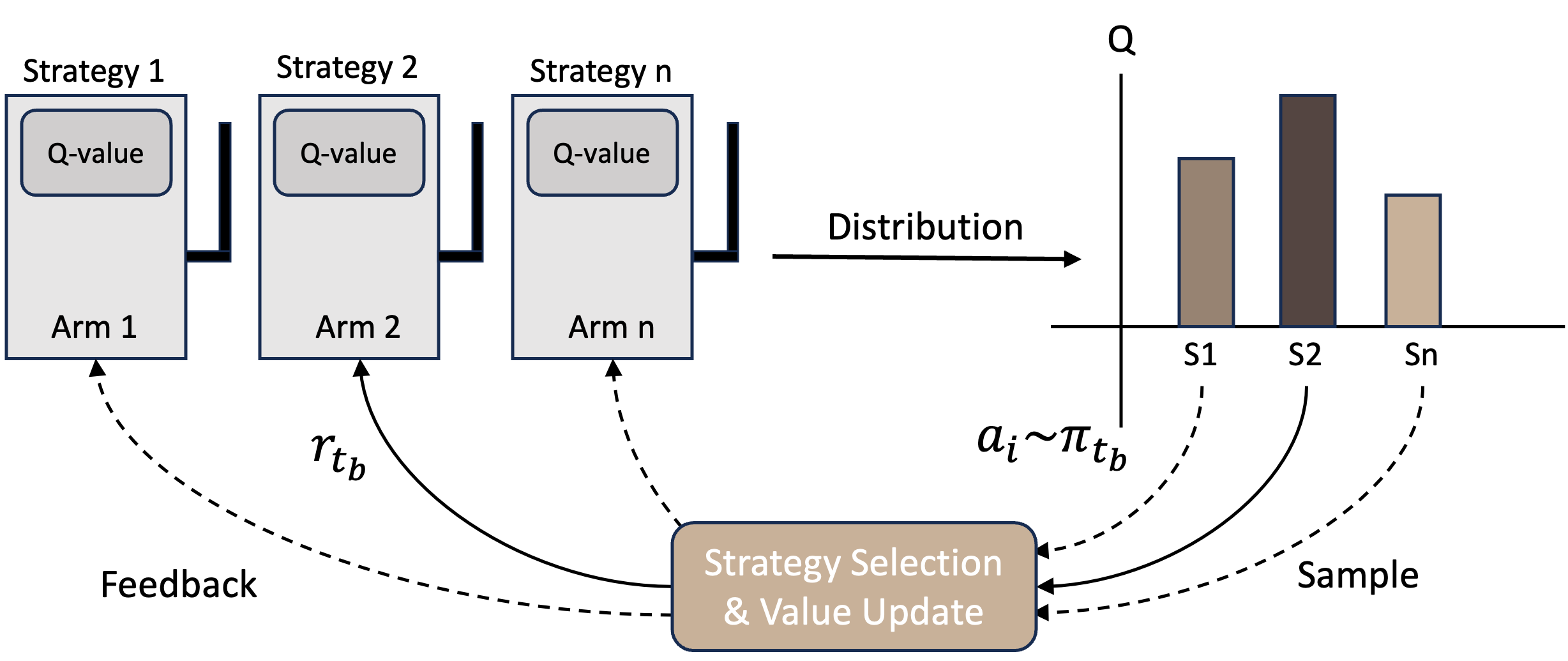}
    \caption{\small \textbf{Dynamic Strategy Optimization with CMAB}. It shows how a CMAB algorithm learns to choose the best strategy. It starts with an initial probability distribution for each strategy. The sample process then selects a strategy $\alpha_i$, and then uses the resulting success or failure feedback $r_{t_b}$ to update both the estimated value (Q-value) of each strategy and the future selection probabilities $\pi_{t_b}$. This process runs iteratively to optimize the strategy selection over time.
    }
    \label{fig:cmab}
\end{figure}

\section{Preliminary}
\label{sec:preliminary}

\begin{figure*}[t]
\centering
\includegraphics[width=0.9\linewidth]{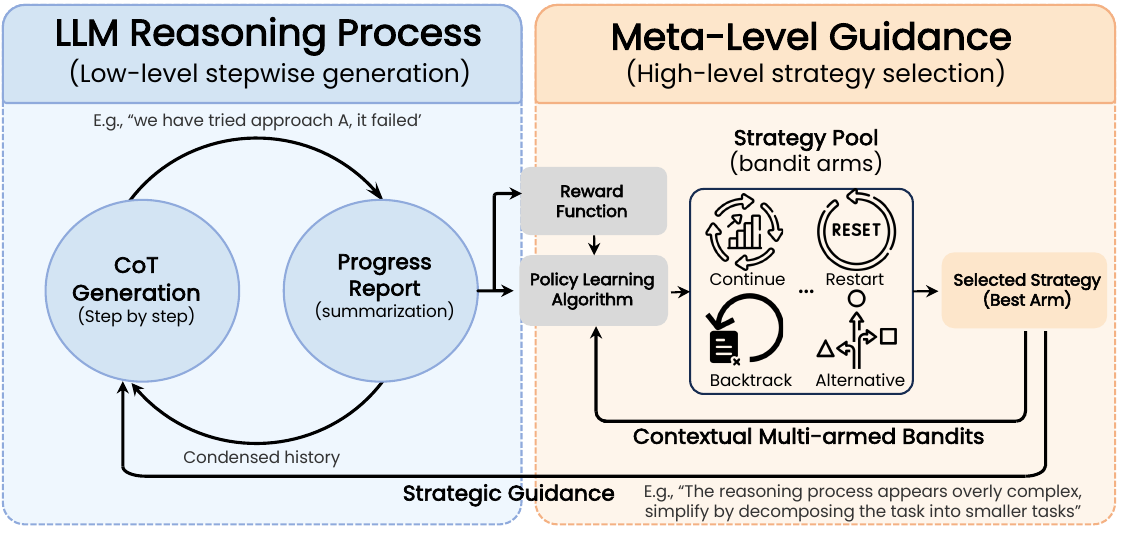}
\caption{\small \textbf{Workflow of \ours{}.} In each round, LLM produces a new reasoning step to extend its reasoning trajectories (\refsec{sec:cot_generation}). The reasoning process is then summarized into a progress report, which provides context for the meta-reasoner (\refsec{sec:progress_report}). Then meta-reasoner employs a CMAB approach to choose a guidance strategy (\refsec{sec:strategy_generation}). The selected strategy then guides the next reasoning step generation, to enable strategic redirection, error correction, or resource optimization. A reward is then computed from the progress report and used to update the bandit algorithm (\refsec{sec:reward_modeling}). The process repeats until the task is complete or the maximum number of rounds is reached.}
\label{fig:illustration}
\end{figure*}

In complex reasoning tasks, \textbf{a key challenge} is deciding the best strategy from several valid options. Consider solving a complex math problem: researchers may employ strategies like decomposition, abstraction, heuristic validation, boundary testing or any other methods at any given step. \textbf{The critical question is \underline{\textit{which}} strategy to use and \underline{\textit{when}}.} This decision making problem aligns well with \emph{contextual multi-armed bandits} (CMABs) problem~\cite{mab_19}, a well-studied framework for making such choices. The framework is designed to balance \emph{exploration} (trying new strategies to learn) and \emph{exploitation} (using strategies known to be effective) based on the current state.

\paragraph{Intuitive Understanding of CMAB.}
Imagine an agent faced with several strategies (called \emph{arms}), and at each step, it observes some information about the current situation (called the \emph{context}). Based on this context, the agent picks one strategy to apply and then receives feedback (a \emph{reward}) indicating how well that strategy performed. The goal of the CMAB problem is to choose strategies over time to maximize the total reward.

As shown in Figure~\ref{fig:cmab}, it illustrates a dynamic strategy selection process using CMAB approach. The agent manages three arms which correspond to three strategies, and each with an associated strategy value (Q-value) that reflects its effectiveness. Based on feedback ($r_{t_b}$) from problem-solving attempts, the agent updates the probability distribution $\pi_{t_b}$, which determines the likelihood of selecting each strategy (e.g., strategy 1 might have a 50\% chance, strategy 2 a 30\% chance, and strategy 3 a 20\% chance). The sample process then picks a specific strategy $ \alpha_i $ (illustrated as the ``chosen strategy") from this distribution to apply to the current problem. The resulting performance feedback is fed back into the controller, which adjusts the Q-value and refines $ \pi_{t_b} $ for the next iteration, optimizing the strategy selection over time.

\paragraph{Formalization of CMABs.}
Formally, at each time step $t$, the model observes a context vector $x_t$ describing the current state and selects an arm $s_t$ from a set of possible strategies $\mathcal{S}$. Then, the model receives a reward $r(s_t, x_t)$ that depends on both the chosen strategy and the context. The model aims to maximize the cumulative reward over $T$ steps:
\begin{equation}
\setlength{\abovedisplayskip}{2pt} 
\setlength{\belowdisplayskip}{2pt} 
    R(T) = \sum_{t=1}^T r(s_t, x_t).
\end{equation}
A popular algorithm for this problem is LinUCB~\cite{li2012contextualbanditapproachpersonalized}, which assumes the expected reward is approximately a linear function of the context. Specifically, for each arm $s$, there is an unknown parameter vector $\theta_s$ such that:
\begin{equation}
\setlength{\abovedisplayskip}{4pt} 
\setlength{\belowdisplayskip}{2pt} 
    \mathbb{E}[r(s, x_t)] \approx x_t^\top \theta_s.
\end{equation}
LinUCB maintains an estimate $\hat{\theta}_s$ of this parameter and a measure of uncertainty about the estimate. The key idea of LinUCB is its approach to the exploration-exploitation dilemma through an \emph{uncertainty bonus}. It quantifies the statistical uncertainty in the reward estimate: arms with higher estimated uncertainty receive a bonus, ensuring they are explored sufficiently.
At each step, LinUCB selects the arm that maximizes a combination of the estimated reward and uncertainty bonus:
\begin{equation}
\setlength{\abovedisplayskip}{4pt} 
\setlength{\belowdisplayskip}{2pt} 
   s_t = \arg\max_{s \in \mathcal{S}} \left[ x_t^\top \hat{\theta}_s + c \sqrt{x_t^\top A_s^{-1} x_t} \right],
\end{equation}
where $A_s$ is a matrix capturing past observations for arm $s$, and $c$ controls how much the model favors exploring uncertain options. The second term encourages trying arms that might perform well but have been less explored.

By using context to guide its choices, LinUCB adapts its strategy selection to the current situation, which aligns with our goal in \ours{}: to dynamically choose the most effective reasoning strategy during inference time. We provide the details about how to leverage these CMABs approaches in our system in Section~\ref{sec:strategy_generation}.

\begin{table*}[t]
\centering
\small
\begin{tabularx}{\linewidth}{@{} X X @{}} 
\toprule
\textbf{Diagnosis} & \textbf{Strategy} \\
\midrule
Progress is insufficient or the current strategy seems ineffective. & 
Restart from scratch and propose alternative strategies. \\
\addlinespace[4pt]
There are mistakes in intermediate steps. & 
Backtrack to the point where the error occurred. \\
\addlinespace[4pt]
The current approach is working well. & 
Continue and provide specific suggestions for the next steps. \\
\midrule
\rowcolor{blue!5}
Ambiguous or conflicting intermediate results are observed. & 
Pause to clarify and disambiguate the current reasoning, then reconcile the discrepancies. \\
\addlinespace[1pt] 
\rowcolor{blue!5}
The reasoning process appears overly complex or convoluted. & 
Simplify by decomposing the task into smaller, manageable sub-tasks. \\
\addlinespace[1pt]
\rowcolor{blue!5}
Evidence of error propagation or low confidence in certain sub-components. & 
Perform targeted verification on critical steps and focus on areas with low confidence. \\
\addlinespace[1pt]
\rowcolor{blue!5}
Repetitive or circular reasoning patterns are detected. & 
Reset to a previously successful checkpoint and explore alternative solution paths. \\
\bottomrule
\end{tabularx}
\caption{\small \textbf{Demonstration} of contextual bandit pairs. The top three rows represent standard strategies, while the \colorbox{blue!5}{\textbf{highlighted bottom four rows}} depict unique strategies generated by Dynamic Contextual Bandits as described in \refsec{method:dynamic_contextual_bandits}.}
\label{tab:contextual_bandit_demonstration}
\end{table*}

\section{Methods}
\label{sec:methods}

Motivated by the intuition that LLMs should concentrate their computational efforts on more promising reasoning paths during inference, we explore two key research questions in this paper: 
\begin{itemize}[leftmargin=*]
\setlength\itemsep{0em}
    \item (1) \textit{How can language models dynamically allocate resources during inference to optimize reasoning and planning?}
    \item (2) \textit{What architectural design enables an effective separation between the reasoning process within the LLM and the meta-level guidance that oversees it?}
\end{itemize}

To address these questions, we propose \textbf{\ours{}}, which endows LLMs with the ability to ``think about how to think". Our framework supervises the LLM's reasoning process and dynamically guides the model to focus on more promising reasoning trajectories during inference time. The framework operates iteratively as illustrated in Figure~\ref{fig:illustration}. At each round $t$, the reasoning process comprises three steps: (1) \textit{CoT generation} by the LLM, (2) \textit{Progress Reporting} to summarize the reasoning progress so far (i.e., this is partly for efficiency, and partly to help the meta-reasoner focus on its main goal of \enquote{advising} rather than being distracted by the details in the CoT), and (3) \textit{Strategy Generation} by the meta-reasoner to optimize subsequent steps. The selection of the strategy almost exactly corresponds to the well-studied problem of contextual multi-armed bandits as illustrated in \refsec{sec:preliminary}. Each strategy can be seen as an arm for the bandit, and the reward of each strategy can be evaluated by the progress of LLM reasoning after applying the strategy. We analogize the process of executing and evaluating each strategy as the act of \enquote{pulling} each arm. The overall goal of our meta-reasoner is to find the best arm (i.e., strategy with highest cumulative rewards) with as few pulls as possible. The complete process of \ours{} is demonstrated in Algorithm~\ref{alg:meta-reasoner}. The prompt for each step is detailed in Appendix \refsec{appendix:prompt_list}.

\subsection{Step-wise CoT Generation}
\label{sec:cot_generation}

In the first step, the LLM generates a reasoning step $s_t$ to extend its reasoning trajectory \(C_t = C_{t-1} \cup \{s_t\}\) based on the user query. This step builds on the previous reasoning history \(C_{t-1}\) and incorporates strategic guidance \(G_{t-1}\) from the meta-reasoner in the prior round. By maintaining the complete reasoning trajectory step-by-step, the model maintains a coherent foundation for refining the progress of reasoning process. This step resembles the long-term reasoning process observed in models like \texttt{o1} and \texttt{r1}, which generate extended CoTs during inference. The key difference is that \ours{} actively intervenes in the reasoning process by providing strategical guidance $G_{t-1}$, enabling the LLMs ``think about how to think". 

\subsection{Progress Reporting}
\label{sec:progress_report}

After updating the CoT, we apply a summarization function $f:C_t \rightarrow P_t$ to distill the most recent $t$ steps of the CoT into a concise progress report $P_t$. This report encapsulates critical elements of the reasoning trajectory, including the degree of progress toward the task objective, the logical consistency of the reasoning steps, and significant milestones or updates achieved. The function $f(\cdot)$ is designed to be computationally efficient, producing a compact summary that enables the meta-reasoner to assess \textbf{high-level} progress without processing the \textbf{granular details} of each step in $C_t$. While this summarization is heuristic in nature, we observe that it may trigger LLMs' capacity for higher-order reasoning by including only essential information in $P_t$, which informs the meta-reasoner's strategic guidance $G_t$. Empirically, this approach fosters more insightful and adaptive strategies, as the meta-reasoner can focus on critical patterns rather than exhaustive step-by-step details. We provide the detailed experiments in \refsec{exp:ablation_study}.

\begin{algorithm}[ht]
\small 
\caption{\small Meta-Reasoning with Contextual Multi-Armed Bandits}
\label{alg:meta-reasoner}
\begin{algorithmic}[1]
\REQUIRE LLM $M$, Bandits $\mathcal{B}$, Initial strategy set $\mathcal{A}_1$, Maximum rounds $T$
\ENSURE Final answer $A_{\mathrm{final}}$
\STATE $C_0 \leftarrow \emptyset$; \quad $\mathcal{B}.\mathrm{Initialize}(\mathcal{A}_1)$
\STATE $G_0 \leftarrow$ default strategy
\FOR{$t = 1$ to $T$}
    \IF{$t > 1$}
        \STATE $P_{t-1} \leftarrow f(C_{t-1})$
        \STATE $x_{t-1} \leftarrow \mathrm{FeatureExtract}(P_{t-1})$
        \STATE \textbf{(Optional)}: $\mathcal{A}_t \leftarrow \mathcal{A}_{t-1} \cup \{\text{new strategies}\}$
        \STATE $a_{t-1} \leftarrow \arg\max_{a \in \mathcal{A}_t} \mathrm{Score}_{\mathcal{B}}(x_{t-1}, a)$
        \STATE $G_t \leftarrow a_{t-1}$
    \ELSE
        \STATE $G_t \leftarrow G_0$
    \ENDIF
    \STATE $s_t \leftarrow M(C_{t-1}, G_t)$
    \STATE $C_t \leftarrow C_{t-1} \cup \{s_t\}$
    \STATE $r_t \leftarrow \mathrm{ComputeReward}(C_t)$
    \IF{$t > 1$}
        \STATE $\mathcal{B}.\mathrm{Update}(x_{t-1}, a_{t-1}, r_t)$
    \ENDIF
    \IF{termination condition met} 
        \STATE \textbf{break}
    \ENDIF
\ENDFOR
\STATE $A_{\mathrm{final}} \leftarrow \mathrm{ExtractAnswer}(C_t)$
\RETURN $A_{\mathrm{final}}$The 
\end{algorithmic}
\end{algorithm}

\subsection{Meta-reasoner Strategy Generation}
\label{sec:strategy_generation}

In the next step, the meta-reasoner evaluates the progress report $P_t$ and selects an appropriate strategy $G_t$ for LLM reasoning (the complete procedure is detailed in Algorithm~\ref{alg:meta-reasoner}). 
We formulate the generation of strategy as a CMAB problem (as defined in \refsec{sec:preliminary}) and explore two settings: (1) a \textit{fixed-strategy} formulation, where the meta-reasoner selects from a predefined set of strategies using a contextual bandit algorithm; and (2) an \textit{advanced} setting, where the meta-reasoner, implemented as an LLM-based agent, dynamically generates or refines strategies.
In both settings, the meta-reasoner leverages the partial-feedback mechanism of MABs to adaptively select strategies based on a reward function that evaluates the quality of reasoning progress after applying the chosen strategy $G_t$. We demonstrate the contextual bandit pair (diagnosis of the progress report (i.e., context) and the corresponding strategy (i.e., bandit) in Table~\ref{tab:contextual_bandit_demonstration}.

\subsubsection{\textbf{Reward Modeling for Progress Report.}}
\label{sec:reward_modeling}
The primary goal of our evaluation mechanism is to quantify how effectively the model’s current reasoning advances toward the task objective (e.g., solving a complex problem), while also monitoring computational cost to promote efficiency. We define a reward function $R: R: P_t \times G_t \to \mathbb{R}$ that integrates two key components: (1) \textbf{solution progress} $S_p$, measuring correctness and adherence to query's constraints (e.g., $ S_p = w_1 \cdot C_c + w_2 \cdot C_a $, where $ C_c $ is correctness, $ C_a $ is adherence, and $ w_1, w_2 $ are weights); and (2) \textbf{resource usage} $ R_u $, reflecting computational cost (e.g., $ R_u = -\alpha \cdot N_s $, where $ N_s $ is the number of reasoning steps and $ \alpha $ is a cost coefficient). The total reward is computed as $ R(G_t, P_t) = \beta \cdot S_p + (1 - \beta) \cdot R_u $, with $ \beta \in [0, 1] $ balancing the trade-off. This evaluation can be implemented via an LLM-based evaluators or external scoring scripts, producing a cumulative score to update the CMAB algorithm. 
To mitigate potential self-favor bias in LLM-based evaluation~\cite{DBLP:journals/corr/abs-2412-05579,DBLP:journals/corr/abs-2411-15594}, where a model may favor its own outputs. We use different models for solution generation and reward modeling. Unless otherwise specified, we use \texttt{gemini-2.5-flash} as the default reward modeling model. To address potential concerns regarding the choice of hyperparameters in the reward function, we conduct a sensitivity analysis in Appendix \refsec{appendix:sensitivity_analysis}.

\subsubsection{\textbf{Fixed Contextual Bandit.}} In the basic version of our framework, the meta-reasoner is modeled as a single contextual bandit that selects from a \textit{fixed, finite} set of $K$ strategies. These strategies include instructions such as \enquote{continue and provide specific suggestions}, \enquote{restart from scratch}, \enquote{backtrack to the point where the error occurred}, or \enquote{propose alternative methods or perspectives to consider}, as detailed in Table~\ref{tab:contextual_bandit_demonstration}. At each round, the LLM produces a \textit{progress report} $P_t$ summarizing its last several reasoning steps, the meta-reasoner then transforms this progress report into a feature vector $x_t$ using a language model and applies a contextual bandit algorithm (e.g., LinUCB~\cite{li2012contextualbanditapproachpersonalized}) to select the best next strategy $a_t$. The LLM then executes that strategy and we collect the reward $r_t$ for $a_t$ based on the reward function. Through iterative MAB algorithm updating, the MAB algorithm learns to select appropriate strategies conditioned on the recent progress report.

\subsubsection{\textbf{Dynamic Contextual Bandit.}}
\label{method:dynamic_contextual_bandits}
The fixed-arm formulation assumes a static set of strategies $G_0$. In practice, the meta-reasoner may itself be an LLM capable of inventing new strategies over time. To accommodate \textit{dynamic} strategies, we allow the meta-reasoner to propose or refine new strategies at round $t$, which generates an expanding collection of strategies, $G_1 \subseteq G_2 \subseteq \cdots \subseteq G_t$. Each newly introduced strategy becomes an additional arm in the contextual multi-armed bandit framework. To encourage at least some exploration on this new arm, we initialize their bandit parameters with a neutral prior, such as $Q(a)=0$ for arm $a$ to avoid strong biases. We further analyze the stability of these new generated dynamic contextual bandits in Appendix \refsec{sec:statbility_of_dynamic_strategy}, ensuring that the system avoids incorporating random or ineffective strategies by grounding new arms in contextual relevance and iterative feedback.

By explicitly separating low-level content generation (handled by the LLM) from high-level strategy decisions (governed by the meta-reasoner’s bandit), the system can effectively avoid getting stuck or wasting excessive resources on poor reasoning paths. In domains where a predefined set of strategies is sufficient, the fixed-arm formulation can simplify the method deployment. While in more open-ended domains where novel tactics may emerge, dynamic-arm extensions give meta-reasoner more flexibility to evolve.

\begin{table}[t]
\centering
\resizebox{\linewidth}{!}{
\begin{tabular}{lc}
\toprule
\textbf{Method}                   & \textbf{Accuracy (\%)}\\ \midrule
GPT-4o-mini + CoT~\cite{yao2023treethoughtsdeliberate} & 4  \\
GPT-4o-mini + SC-CoT~\cite{yao2023treethoughtsdeliberate} & 9   \\
GPT-4o-mini + IO {\scriptsize(best of 100)}~\cite{yao2023treethoughtsdeliberate} & 33   \\
GPT-4o-mini + CoT {\scriptsize(best of 100)}~\cite{yao2023treethoughtsdeliberate} & 49   \\
Gemini-Exp-1206 + IO {\scriptsize(best of 100)}~\cite{yao2023treethoughtsdeliberate} &  38  \\
Gemini-Exp-1206 + CoT {\scriptsize(best of 100)}~\cite{yao2023treethoughtsdeliberate} &  60   \\
\hdashline\\[-8pt]
GPT-4o-mini + ToT {\scriptsize($b = 1$)}~\cite{yao2023treethoughtsdeliberate} & 32      \\
GPT-4o-mini+ ToT {\scriptsize($b = 5$)}~\cite{yao2023treethoughtsdeliberate} & 65      \\
GPT-4o-mini + Reflexion~\cite{shinn2024reflexion} & 53 \\
GPT-4o-mini + MACM~\cite{lei2024macmutilizingmultiagent} & 80     \\
GPT-4o-mini + Meta-Reasoner {\scriptsize(our work)} & 89 \\
GPT-4o + Meta-Reasoner {\scriptsize(our work)} & 92 \\
Gemini-Exp-1206 + Meta-Reasoner {\scriptsize(our work)} & 94 \\
Qwen3-8B + Meta-Reasoner {\scriptsize(our work)} & 87 \\
DS-R1-Distill-Qwen-14B + Meta-Reasoner {\scriptsize(our work)} & 93 \\
\hdashline\\[-8pt]
o1-mini + IO & 89 \\
o1-preview + IO & 93 \\
\bottomrule
\end{tabular}}
\caption{Performance of different models on 24-points Game~\cite{yao2023treethoughtsdeliberate} ($b$: search breadth)}
\label{tab:24_points_results}
\end{table}

\begin{table}[t]
\centering
\resizebox{\linewidth}{!}{
\begin{tabular}{lc}
\toprule
\textbf{Method}                   & \textbf{Accuracy (\%)} \\ 
\midrule
GPT-4o-mini + CoT & 39.46   \\
Gemini-Exp-1206 + CoT &  43.12  \\
\hdashline\\[-8pt]
GPT-4o-mini + Reflexion~\cite{shinn2024reflexion} & 74.32 \\
GPT-4 Turbo + MACM~\cite{lei2024macmutilizingmultiagent} & 79.41     \\
GPT-4o-mini + Meta-Reasoner {\scriptsize(our work)} & 84.13 \\
Gemini-Exp-1206 + Meta-Reasoner {\scriptsize(our work)} & 86.32 \\
Qwen3-8B + Meta-Reasoner {\scriptsize(our work)} & 82.93 \\
DS-R1-Distill-Qwen-14B + Meta-Reasoner {\scriptsize(our work)} & 87.40 \\
\bottomrule
\end{tabular}}
\caption{Performance of different models on TheoremQA~\cite{chen2023theoremqatheoremdrivenquestion}}
\label{tab:theoremqa_results}
\end{table}

\section{Experiments}
\label{sec:experiments}

Due to page limits, we put all the experiment settings and training details in Appendix \refsec{sec:training_details}. In this section, we present the main results, ablation study, analysis regarding efficiency, rewards accumulation, and qualitative assessment.

\subsection{Main Results}
\label{exp:main_results}
    
We compare the accuracy of different prompting methods across different backbone models on SciBench (as shown in Table~\ref{tab:sci_bench_results}), 24-points game (as shown in Table~\ref{tab:24_points_results}) and TheoremQA (as shown in Table~\ref{tab:theoremqa_results}). 
We find that basic prompting strategies, such as CoT and SC-CoT, show limited effectiveness, achieving only 4\% and 9\% accuracy on 24-point games, respectively. Incorporating IO strategy with \enquote{Best of 100} samples improves accuracy to 33\%, but it remains far behind advanced methods like MACM or ToT. Strategies like ToT illustrate the performance of exploring broader reasoning paths in a structured manner, with accuracy increasing from 45\% to 74\%. The more advanced iterative methods like Reflexion and MACM further demonstrate the value of refined reasoning frameworks incorporating multi-step reflection and error checking. Our proposed \ours{} outperforms these approaches, achieving 89\% accuracy with GPT-4o-mini and 92\% with GPT-4o, showcasing its ability to dynamically guide reasoning, correct errors, and focus resources effectively. Compared to specialized models like o1-mini, our method equipped with much cheaper models like GPT-4o-mini delivers comparable performance.


Overall, the Meta-Reasoner framework provides a compatible approach to reasoning-intensive tasks, achieving high accuracy with dynamic and efficient problem-solving strategies. The results on SciBench (Table~\ref{tab:sci_bench_results}) and TheoremQA (Table~\ref{tab:theoremqa_results}) also demonstrate similar findings and show that \ours{} generally achieves better performance compared to the baselines and the results are consistent across different models.

\begin{table}[t]
\centering
\resizebox{\linewidth}{!}{
\begin{tabular}{lccc}
\toprule
\textbf{Method} & \textbf{Diff(\%)} & \textbf{Stat(\%)} & \textbf{Calc(\%)} \\
\midrule
Phi-4 + CoT & 17.42 & 28.42 & 32.93 \\
Llama-3.1-instruct + CoT & 33.14 & 49.72 & 54.18\\
Gemini-Exp-1206 + CoT & 36.32    & 56.73 & 59.24 \\
Gemini-Exp-1206 + SC-CoT & 38.73    & 59.12 & 64.11 \\
GPT-4o-mini + CoT  & 33.12    & 55.71 & 58.10 \\
GPT-4o-mini + SC-CoT       & 37.33    & 56.67 & 63.81 \\
GPT-4o-mini + MCR       & 40.12    & 58.21 & 67.42 \\
\hdashline\\[-8pt]
GPT-4o-mini + MACM~\cite{lei2024macmutilizingmultiagent} & 54.78 & 67.13 & 65.77 \\
GPT-4o + MACM~\cite{lei2024macmutilizingmultiagent} & 61.42 & 78.32 & 76.72 \\
GPT-4o-mini + Meta-Reasoner {\scriptsize(our work)} & 60.32 & 73.64 & 80.23 \\
GPT-4o + Meta-Reasoner {\scriptsize(our work)} & 67.14 & 83.29 &  84.17 \\
Qwen3-8B + Meta-Reasoner {\scriptsize(our work)} & 59.31 & 71.80 & 78.65 \\
DS-R1-Distill-Qwen-14B + Meta-Reasoner {\scriptsize(our work)} & 66.70 & 78.43 & 84.55 \\
\bottomrule
\end{tabular}}
\caption{Performance of different models on SciBench Math Subsets~\cite{wang2024scibenchevaluatingcollegelevel}}
\label{tab:sci_bench_results}
\end{table}

\begin{table}[t]
\centering
\resizebox{\linewidth}{!}{
\begin{tabular}{llcc}
\toprule
\textbf{Method} & \textbf{Base Model} & \textbf{MATH-500 (\%)} & \textbf{AIME 2024 (Pass@1)} \\
\midrule
rStar-Math~\cite{guan2025rstar0math0} & Qwen2.5-Math-7B & 90.0 & 53.3 \\
rStar-Math~\cite{guan2025rstar0math0} & Phi3-mini-3.8B & 86.4 & 43.3 \\
\midrule
MCTS-RAP~\cite{hao2023reasoning} & GPT-4o-mini & 72.5 & 14.5 \\
MCTS-RAP~\cite{hao2023reasoning} & GPT-4o & 79.0 & 23.4 \\
MCTS-RAP~\cite{hao2023reasoning} & DeepSeek-R1-Distill-Qwen-7B & 84.3 & 40.2 \\
\midrule
\textbf{Meta-Reasoner (Ours)} & GPT-4o-mini & 85.6 & 26.7 \\
\textbf{Meta-Reasoner (Ours)} & GPT-4o & 87.3 & 33.3 \\
\textbf{Meta-Reasoner (Ours)} & DeepSeek-R1-Distill-Qwen-7B & \textbf{92.3} & \textbf{55.5} \\
\bottomrule
\end{tabular}}
\caption{\small Performance on Math-500~\cite{lightman2023letsverifystep} and AIME-2024~\cite{guan2025rstar0math0,aimo2024validationaime}.}
\label{tab:high_difficulty_benchmarks}
\end{table}


\subsection{Ablation Study}
\label{exp:ablation_study}

We further conduct an ablation study to analyze each component contribution of \ours{}. Specifically, we consider the following setup: (1) w/o progress report: we replace the progress reporting process with directly considering the entire CoT history without summarization; (2) w/o MAB: instead of using MAB to select the proper strategy, we directly prompting an LLM to provide the reasoning strategy.

\begin{table}[t]
    \centering
    \resizebox{\linewidth}{!}{
    \begin{tabular}{llccc}
    \toprule
    \textbf{Model} &   \textbf{Variant}  & \textbf{Game-of-24} & \textbf{TheoremQA} \\
    \midrule
    \multirow{4}{*}{GPT-4o-mini} & 
        Full Method & 89 &  84.13 \\
    &    w/o Progress Report & 85 & 79.42 \\
    &   w/o MAB (direct arm selection) & 82 & 80.74\\
    &   w/o MAB (CoT) & 4 & 39.46  \\
    \midrule
    \multirow{4}{*}{Gemini-Exp-1206} & 
        Full Method & 94 & 86.32   \\
    &    w/o Progress Report & 91 & 81.78  \\
    &   w/o MAB (direct arm selection) & 87 & 82.14\\
    &   w/o MAB (CoT)  & 11 & 43.12 \\
    \bottomrule
    \end{tabular}}
        \caption{\small Ablation study of \ours{}. Direct arm selection refers to prompting LLM to directly select a strategy based on recent progress report.}
    \label{tab:ablation_study}
\end{table}

In Table~\ref{tab:ablation_study}, we show that when removing progress reporting (\enquote{w/o Progress Report}), the overall performance moderately degrades and we hypothesize it is due to the concise intermediate summarizations can help \ours{} only consider the high-level strategy instead of being confused with too many details of the reasoning process. We also find that removing the MAB brings a more pronounced effect, especially when strategy selection falls back to a direct chain-of-thought approach (\enquote{w/o MAB (CoT)}). It verifies the effect of our meta-reasoner module to help the model stay on track for getting an optimal solution. In
Table~\ref{tab:fixed_dynamic}, we compare fixed and dynamic bandit variants on the game of 24 and theoremQA. We find that using a fixed set of strategies (e.g., $K=3$ and $K=5$) yields lower performance compared to the dynamic approach which adaptively explores more strategies (shown by larger unique strategies). The results highlight the benefit of flexibly allocating diverse reasoning strategies using LLM in-context learning capabilities.

\subsection{Analysis}
\label{sec:analysis}

\begin{figure*}[t]
    \centering
    \includegraphics[width=\linewidth]{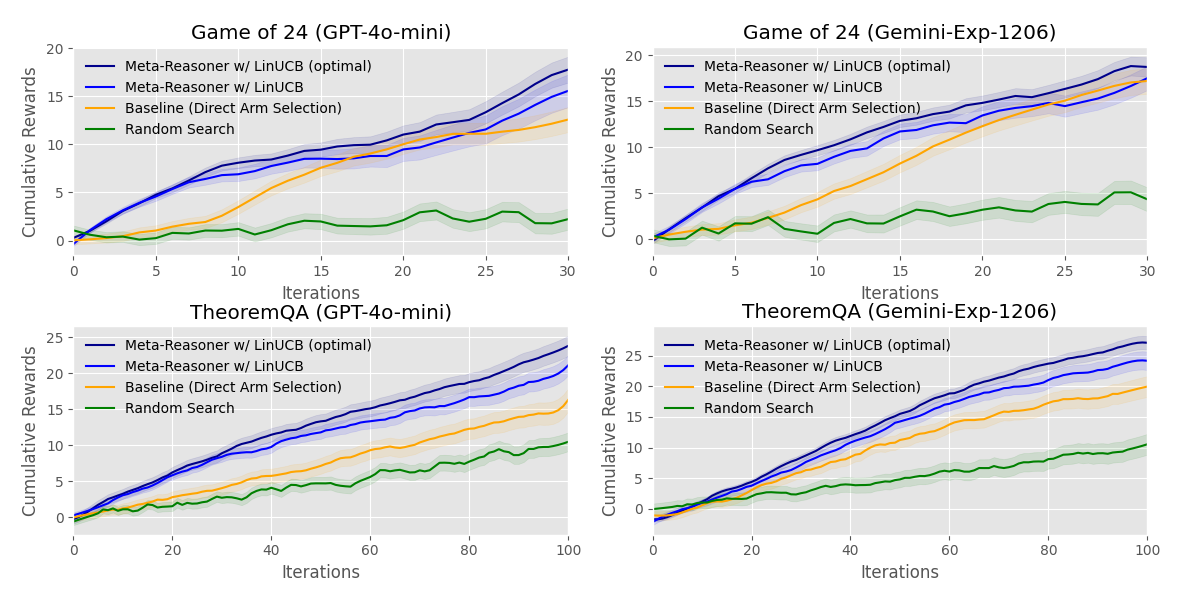}
    \caption{\small Cumulative reward of different settings across iteration. We compare our method using LinUCB with baseline (direct arm selection), and random search methods across two tasks—Game of 24 (top row) and TheoremQA (bottom row) using GPT-4o-mini (left) and Gemini-Exp-1206 (right).}
    \label{fig:accumulated_rewards}
\end{figure*}

\paragraph{Effectiveness of Meta-reasoner.}
Figure~\ref{fig:accumulated_rewards} demonstrates the cumulative rewards across iterations. 
We compare our MAB-based approach with a baseline that directly prompts an LLM to select an arm (or \enquote{strategy}), referred to as \textit{Baseline (Direct Arm Selection)}; the prompt details are in Appendix \refsec{appendix:prompt_list}. Results show that the MAB-based meta-reasoner (using LinUCB~\cite{li2012contextualbanditapproachpersonalized}) consistently outperforms both direct LLM decision-making and random search across two tasks (Game of 24 and TheoremQA) and two model scales (GPT-4o-mini and Gemini-Exp-1206). While direct LLM prompting yields reasonable initial performance and random search requires minimal setup, neither approach effectively balances exploration and exploitation. In contrast, the MAB updating strategy leverages feedback from prior iterations to adaptively refine action selection (e.g., choosing an appropriate strategy based on CoT reasoning), steadily increasing cumulative rewards.

\begin{table}[ht]
    \centering
    \resizebox{\linewidth}{!}{
    \begin{tabular}{ccccc}
    \toprule
       Bandit Type  & Game-of-24(\%) & \#US & TheoremQA(\%) & \#US \\
    \midrule
        Fixed (K=3) & 65 & 3 & 72.34 & 3 \\
        Fixed (K=5) & 72 & 5 & 79.17 & 5 \\
        Dynamic & 89 & 14 & 84.13 & 21 \\
        \bottomrule
    \end{tabular}}
        \caption{\small Fixed vs. Dynamic Bandit Variants over \texttt{GPT-4o-mini}. \#US: Number of Unique Strategies.}
    \label{tab:fixed_dynamic}
\end{table}

\begin{figure}[ht]
     \centering
    \begin{subfigure}[b]{0.49\linewidth}
         \centering
        \includegraphics[width=1\textwidth,trim={0cm 0.2cm 0cm 0cm}]{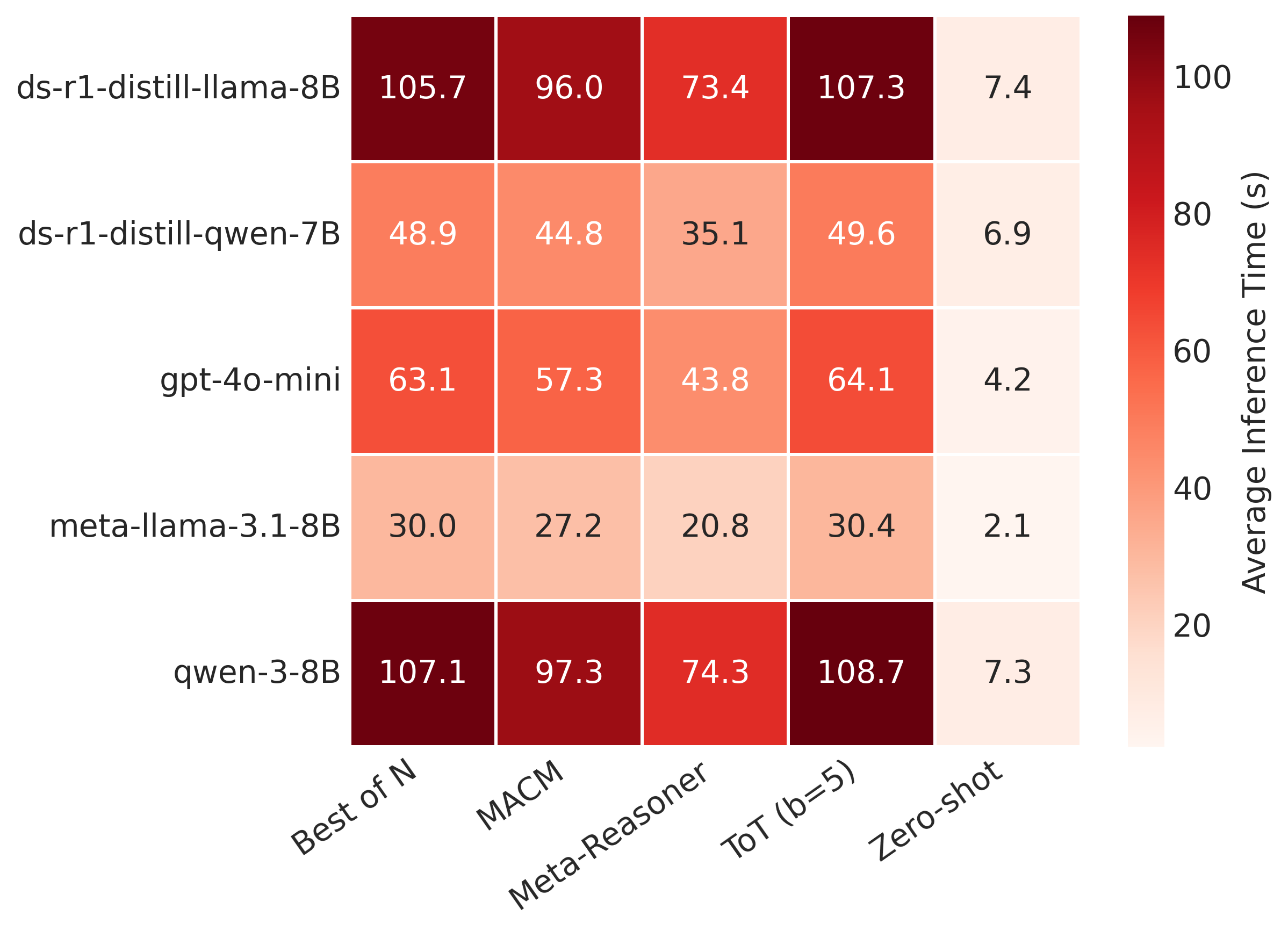}
         \caption{\small Game of 24 Task.}
        \label{fig:inference_time_heatmap_game_24}
     \end{subfigure}
     \hfill
     \hfill
     \hfill
     \hfill 
     \begin{subfigure}[b]{0.49\linewidth}
         \centering
        \includegraphics[width=1\textwidth,trim={0cm 0.2cm 0cm 0cm}]{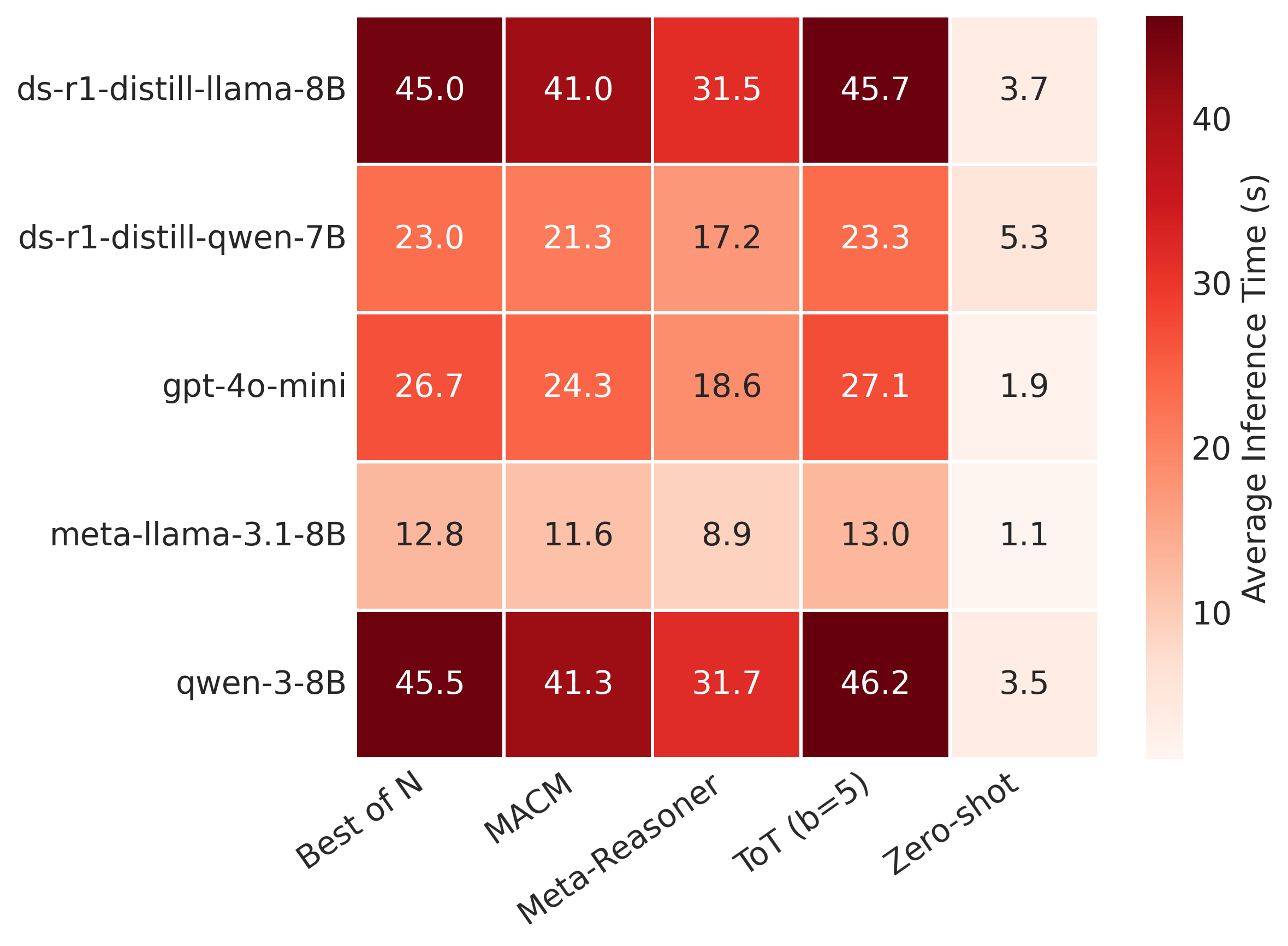}
         \caption{\small TheoremQA Task}
        \label{fig:inference_heatmap_theoremqa}
     \end{subfigure}
     \caption{Inference time heatmap comparison.}
     \label{fig:inference_time_heatmap}
\end{figure}

\paragraph{Inference Efficiency.} We measure the inference efficiency of our proposed method. In Figure~\ref{fig:inference_time_heatmap}, we calculate the average inference time on different tasks across different models and methods. It demonstrates Meta-Reasoner's significant advantages over existing reasoning frameworks across multiple benchmarks. Compared to o1-preview in the zero-shot setting, our method achieved roughly 51-55\% inference time reduction while maintaining comparable performance. Among reasoning methods using identical base models, Meta-Reasoner consistently outperformed alternatives: requiring 29-33\% less inference time than ToT~\cite{yao2023treethoughtsdeliberate}, Reflexion~\cite{shinn2024reflexion}, MACM~\cite{lei2024macmutilizingmultiagent} and Best-of-N techniques.We further provide the demonstration of the token usage comparison in Figure~\ref{fig:token_usage}. The efficiency results positions \ours{} as a scalable solution for complex reasoning tasks where computational resources are constrained.

\paragraph{Iterative Reasoning and Memory Management.}
Unlike traditional CoT methods that generate reasoning process in a single pass, \ours{} pauses generation periodically to evaluate the reasoning progress and adaptively switch the reasoning strategies using a multi-armed bandit-based algorithm. This iterative design adds additional memory overhead due to maintaining context across multiple LLM calls. To mitigate this, we summarize the entire reasoning history into compact progress reports (\refsec{sec:progress_report}), greatly reducing the token load per iteration. As demonstrated in Figure~\ref{fig:inference_time_heatmap}, the savings achieved by avoiding unproductive reasoning paths significantly outweigh the modest overhead introduced by the strategy selection process.

\paragraph{Generalizability of \ours{}.}
We test the generalizability of \ours{} on creative writing task, which differs from math and logical reasoning. Following ToT~\cite{yao2023treethoughtsdeliberate}, the input consists of four random sentences, and the model must generate a four-paragraph story that ends with those sentences. We use the same automatic evaluation protocol as ToT: a GPT-4 zero-shot judge assigns a scalar coherence score from 1 to 10 for each output. Table~\ref{tab:results_on_creative_writing} shows that Meta-Reasoner achieves a coherence score of 7.68. This is competitive with ToT + Prompt Refined (7.91) and clearly better than IO (6.19) and CoT (6.93). These results suggest that \ours{} is domain-agnostic, it can transfer beyond mathematical objectives and remains effective in open-ended creative generation.

\begin{table}[ht]
    \centering
    \resizebox{\linewidth}{!}{
    \begin{tabular}{lc}
    \toprule
        \textbf{Method} & \textbf{GPT-4 Coherence Score} \\ 
    \midrule
        IO (Input-Output) & 6.19 \\
        IO + Prompt Refined (\(k = 5\)) & 7.67 \\
        CoT (Chain-of-Thought)~\cite{wei2022chain} & 6.93 \\
        \hdashline\\[-8pt]
        ToT (Tree-of-Thoughts)~\cite{yao2023treethoughtsdeliberate} & 7.56 \\
        ToT + Prompt Refined~\cite{yao2023treethoughtsdeliberate} & 7.91 \\
        \textbf{\ours{}} & 7.68 \\
    \bottomrule
    \end{tabular}}
    \caption{Generalizability of \ours{} on the creative writing task~\cite{yao2023treethoughtsdeliberate}. Higher GPT-4 coherence scores are better.}
    \label{tab:results_on_creative_writing}
\end{table}

\section{Conclusion}

In this work, we introduce \ours{}, a framework designed to enhance the reasoning capabilities of LLMs and optimize the inference-time efficiency. By operating as an \enquote{advisor}, meta-reasoner dynamically evaluates the reasoning process and provides high-level strategic guidance, addressing key limitations of long CoT reasoning, such as compounding errors and inefficiency in inference computing. Unlike conventional approaches, \ours{} focuses on global oversight rather than granular step-by-step reasoning, enabling LLMs to avoid unproductive lines of thought and better allocate computational resources. Experiments highlight the potential of dynamic reasoning to overcome inherent challenges in the LLM reasoning and also show promise in broader applications, offering a scalable and adaptable solution for reasoning-intensive tasks.

\section*{Limitations}


Our proposed Meta-Reasoner framework, while effective at enhancing inference-time reasoning, remains limited to text-based problems and struggles to address tasks requiring other modalities, such as geometry. Overcoming these challenges calls for further advancements in the model’s cognitive capabilities.

\bibliography{ref}

@article{zhang2025survey,
  title={A Survey on Test-Time Scaling in Large Language Models: What, How, Where, and How Well?},
  author={Zhang, Qiyuan and Lyu, Fuyuan and Sun, Zexu and Wang, Lei and Zhang, Weixu and Hua, Wenyue and Wu, Haolun and Guo, Zhihan and Wang, Yufei and Muennighoff, Niklas and others},
  journal={arXiv preprint arXiv:2503.24235},
  year={2025}
}

@article{guan2025rstar0math0,
  title   = {rStar-Math: Small LLMs Can Master Math Reasoning with Self-Evolved Deep Thinking},
  author  = {Xinyu Guan and Li Lyna Zhang and Yifei Liu and Ning Shang and Youran Sun and Yi Zhu and Fan Yang and Mao Yang},
  year    = {2025},
  journal = {arXiv preprint arXiv: 2501.04519}
}

@article{deepseek-ai2025deepseekr1,
  title   = {DeepSeek-R1: Incentivizing Reasoning Capability in LLMs via Reinforcement Learning},
  author  = {DeepSeek-AI and Daya Guo and Dejian Yang and Haowei Zhang and Junxiao Song and Ruoyu Zhang and Runxin Xu and Qihao Zhu and Shirong Ma and Peiyi Wang and Xiao Bi and Xiaokang Zhang and Xingkai Yu and Yu Wu and Z. F. Wu and Zhibin Gou and Zhihong Shao and Zhuoshu Li and Ziyi Gao and Aixin Liu and Bing Xue and Bingxuan Wang and Bochao Wu and Bei Feng and Chengda Lu and Chenggang Zhao and Chengqi Deng and Chenyu Zhang and Chong Ruan and Damai Dai and Deli Chen and Dongjie Ji and Erhang Li and Fangyun Lin and Fucong Dai and Fuli Luo and Guangbo Hao and Guanting Chen and Guowei Li and H. Zhang and Han Bao and Hanwei Xu and Haocheng Wang and Honghui Ding and Huajian Xin and Huazuo Gao and Hui Qu and Hui Li and Jianzhong Guo and Jiashi Li and Jiawei Wang and Jingchang Chen and Jingyang Yuan and Junjie Qiu and Junlong Li and J. L. Cai and Jiaqi Ni and Jian Liang and Jin Chen and Kai Dong and Kai Hu and Kaige Gao and Kang Guan and Kexin Huang and Kuai Yu and Lean Wang and Lecong Zhang and Liang Zhao and Litong Wang and Liyue Zhang and Lei Xu and Leyi Xia and Mingchuan Zhang and Minghua Zhang and Minghui Tang and Meng Li and Miaojun Wang and Mingming Li and Ning Tian and Panpan Huang and Peng Zhang and Qiancheng Wang and Qinyu Chen and Qiushi Du and Ruiqi Ge and Ruisong Zhang and Ruizhe Pan and Runji Wang and R. J. Chen and R. L. Jin and Ruyi Chen and Shanghao Lu and Shangyan Zhou and Shanhuang Chen and Shengfeng Ye and Shiyu Wang and Shuiping Yu and Shunfeng Zhou and Shuting Pan and S. S. Li and Shuang Zhou and Shaoqing Wu and Shengfeng Ye and Tao Yun and Tian Pei and Tianyu Sun and T. Wang and Wangding Zeng and Wanjia Zhao and Wen Liu and Wenfeng Liang and Wenjun Gao and Wenqin Yu and Wentao Zhang and W. L. Xiao and Wei An and Xiaodong Liu and Xiaohan Wang and Xiaokang Chen and Xiaotao Nie and Xin Cheng and Xin Liu and Xin Xie and Xingchao Liu and Xinyu Yang and Xinyuan Li and Xuecheng Su and Xuheng Lin and X. Q. Li and Xiangyue Jin and Xiaojin Shen and Xiaosha Chen and Xiaowen Sun and Xiaoxiang Wang and Xinnan Song and Xinyi Zhou and Xianzu Wang and Xinxia Shan and Y. K. Li and Y. Q. Wang and Y. X. Wei and Yang Zhang and Yanhong Xu and Yao Li and Yao Zhao and Yaofeng Sun and Yaohui Wang and Yi Yu and Yichao Zhang and Yifan Shi and Yiliang Xiong and Ying He and Yishi Piao and Yisong Wang and Yixuan Tan and Yiyang Ma and Yiyuan Liu and Yongqiang Guo and Yuan Ou and Yuduan Wang and Yue Gong and Yuheng Zou and Yujia He and Yunfan Xiong and Yuxiang Luo and Yuxiang You and Yuxuan Liu and Yuyang Zhou and Y. X. Zhu and Yanhong Xu and Yanping Huang and Yaohui Li and Yi Zheng and Yuchen Zhu and Yunxian Ma and Ying Tang and Yukun Zha and Yuting Yan and Z. Z. Ren and Zehui Ren and Zhangli Sha and Zhe Fu and Zhean Xu and Zhenda Xie and Zhengyan Zhang and Zhewen Hao and Zhicheng Ma and Zhigang Yan and Zhiyu Wu and Zihui Gu and Zijia Zhu and Zijun Liu and Zilin Li and Ziwei Xie and Ziyang Song and Zizheng Pan and Zhen Huang and Zhipeng Xu and Zhongyu Zhang and Zhen Zhang},
  year    = {2025},
  journal = {arXiv preprint arXiv: 2501.12948}
}

@article{zhang2024learning,
  title={Learning to check: Unleashing potentials for self-correction in large language models},
  author={Zhang, Che and Xiao, Zhenyang and Han, Chengcheng and Lian, Yixin and Fang, Yuejian},
  journal={arXiv preprint arXiv:2402.13035},
  year={2024}
}

@article{openai2024openai,
  title   = {OpenAI o1 System Card},
  author  = {OpenAI and : and Aaron Jaech and Adam Kalai and Adam Lerer and Adam Richardson and Ahmed El-Kishky and Aiden Low and Alec Helyar and Aleksander Madry and Alex Beutel and Alex Carney and Alex Iftimie and Alex Karpenko and Alex Tachard Passos and Alexander Neitz and Alexander Prokofiev and Alexander Wei and Allison Tam and Ally Bennett and Ananya Kumar and Andre Saraiva and Andrea Vallone and Andrew Duberstein and Andrew Kondrich and Andrey Mishchenko and Andy Applebaum and Angela Jiang and Ashvin Nair and Barret Zoph and Behrooz Ghorbani and Ben Rossen and Benjamin Sokolowsky and Boaz Barak and Bob McGrew and Borys Minaiev and Botao Hao and Bowen Baker and Brandon Houghton and Brandon McKinzie and Brydon Eastman and Camillo Lugaresi and Cary Bassin and Cary Hudson and Chak Ming Li and Charles de Bourcy and Chelsea Voss and Chen Shen and Chong Zhang and Chris Koch and Chris Orsinger and Christopher Hesse and Claudia Fischer and Clive Chan and Dan Roberts and Daniel Kappler and Daniel Levy and Daniel Selsam and David Dohan and David Farhi and David Mely and David Robinson and Dimitris Tsipras and Doug Li and Dragos Oprica and Eben Freeman and Eddie Zhang and Edmund Wong and Elizabeth Proehl and Enoch Cheung and Eric Mitchell and Eric Wallace and Erik Ritter and Evan Mays and Fan Wang and Felipe Petroski Such and Filippo Raso and Florencia Leoni and Foivos Tsimpourlas and Francis Song and Fred von Lohmann and Freddie Sulit and Geoff Salmon and Giambattista Parascandolo and Gildas Chabot and Grace Zhao and Greg Brockman and Guillaume Leclerc and Hadi Salman and Haiming Bao and Hao Sheng and Hart Andrin and Hessam Bagherinezhad and Hongyu Ren and Hunter Lightman and Hyung Won Chung and Ian Kivlichan and Ian O'Connell and Ian Osband and Ignasi Clavera Gilaberte and Ilge Akkaya and Ilya Kostrikov and Ilya Sutskever and Irina Kofman and Jakub Pachocki and James Lennon and Jason Wei and Jean Harb and Jerry Twore and Jiacheng Feng and Jiahui Yu and Jiayi Weng and Jie Tang and Jieqi Yu and Joaquin Quiñonero Candela and Joe Palermo and Joel Parish and Johannes Heidecke and John Hallman and John Rizzo and Jonathan Gordon and Jonathan Uesato and Jonathan Ward and Joost Huizinga and Julie Wang and Kai Chen and Kai Xiao and Karan Singhal and Karina Nguyen and Karl Cobbe and Katy Shi and Kayla Wood and Kendra Rimbach and Keren Gu-Lemberg and Kevin Liu and Kevin Lu and Kevin Stone and Kevin Yu and Lama Ahmad and Lauren Yang and Leo Liu and Leon Maksin and Leyton Ho and Liam Fedus and Lilian Weng and Linden Li and Lindsay McCallum and Lindsey Held and Lorenz Kuhn and Lukas Kondraciuk and Lukasz Kaiser and Luke Metz and Madelaine Boyd and Maja Trebacz and Manas Joglekar and Mark Chen and Marko Tintor and Mason Meyer and Matt Jones and Matt Kaufer and Max Schwarzer and Meghan Shah and Mehmet Yatbaz and Melody Y. Guan and Mengyuan Xu and Mengyuan Yan and Mia Glaese and Mianna Chen and Michael Lampe and Michael Malek and Michele Wang and Michelle Fradin and Mike McClay and Mikhail Pavlov and Miles Wang and Mingxuan Wang and Mira Murati and Mo Bavarian and Mostafa Rohaninejad and Nat McAleese and Neil Chowdhury and Neil Chowdhury and Nick Ryder and Nikolas Tezak and Noam Brown and Ofir Nachum and Oleg Boiko and Oleg Murk and Olivia Watkins and Patrick Chao and Paul Ashbourne and Pavel Izmailov and Peter Zhokhov and Rachel Dias and Rahul Arora and Randall Lin and Rapha Gontijo Lopes and Raz Gaon and Reah Miyara and Reimar Leike and Renny Hwang and Rhythm Garg and Robin Brown and Roshan James and Rui Shu and Ryan Cheu and Ryan Greene and Saachi Jain and Sam Altman and Sam Toizer and Sam Toyer and Samuel Miserendino and Sandhini Agarwal and Santiago Hernandez and Sasha Baker and Scott McKinney and Scottie Yan and Shengjia Zhao and Shengli Hu and Shibani Santurkar and Shraman Ray Chaudhuri and Shuyuan Zhang and Siyuan Fu and Spencer Papay and Steph Lin and Suchir Balaji and Suvansh Sanjeev and Szymon Sidor and Tal Broda and Aidan Clark and Tao Wang and Taylor Gordon and Ted Sanders and Tejal Patwardhan and Thibault Sottiaux and Thomas Degry and Thomas Dimson and Tianhao Zheng and Timur Garipov and Tom Stasi and Trapit Bansal and Trevor Creech and Troy Peterson and Tyna Eloundou and Valerie Qi and Vineet Kosaraju and Vinnie Monaco and Vitchyr Pong and Vlad Fomenko and Weiyi Zheng and Wenda Zhou and Wes McCabe and Wojciech Zaremba and Yann Dubois and Yinghai Lu and Yining Chen and Young Cha and Yu Bai and Yuchen He and Yuchen Zhang and Yunyun Wang and Zheng Shao and Zhuohan Li},
  year    = {2024},
  journal = {arXiv preprint arXiv: 2412.16720}
}

@article{yao2024mulberry0,
  title   = {Mulberry: Empowering MLLM with o1-like Reasoning and Reflection via Collective Monte Carlo Tree Search},
  author  = {Huanjin Yao and Jiaxing Huang and Wenhao Wu and Jingyi Zhang and Yibo Wang and Shunyu Liu and Yingjie Wang and Yuxin Song and Haocheng Feng and Li Shen and Dacheng Tao},
  year    = {2024},
  journal = {arXiv preprint arXiv: 2412.18319}
}

@inproceedings{hao2023reasoning,
  title={Reasoning with language model is planning with world model},
  author={Hao, Shibo and Gu, Yi and Ma, Haodi and Hong, Joshua and Wang, Zhen and Wang, Daisy and Hu, Zhiting},
  booktitle={Proceedings of the 2023 Conference on Empirical Methods in Natural Language Processing},
  pages={8154--8173},
  year={2023}
}

@article{wu2024beyond,
  title   = {Beyond Examples: High-level Automated Reasoning Paradigm in In-Context Learning via MCTS},
  author  = {Jinyang Wu and Mingkuan Feng and Shuai Zhang and Feihu Che and Zengqi Wen and Chonghua Liao and Jianhua Tao},
  year    = {2024},
  journal = {arXiv preprint arXiv: 2411.18478}
}

@misc{aimo2024validationaime,
  author       = {{AI-MO Team}},
  title        = {AIMO Validation AIME Dataset},
  year         = {2024},
  publisher    = {Hugging Face},
  howpublished = {\url{https://huggingface.co/datasets/AI-MO/aimo-validation-aime}},
  note         = {Dataset containing problems from AIME 2022-2024}
}

@article{guo2023evoprompt0,
  title   = {EvoPrompt: Connecting LLMs with Evolutionary Algorithms Yields Powerful Prompt Optimizers},
  author  = {Qingyan Guo and Rui Wang and Junliang Guo and Bei Li and Kaitao Song and Xu Tan and Guoqing Liu and Jiang Bian and Yujiu Yang},
  year    = {2023},
  journal = {arXiv preprint arXiv: 2309.08532}
}

@article{besta2023graphthoughtssolving,
  title = {Graph of Thoughts: {{Solving}} Elaborate Problems with Large Language Models},
  author = {Besta, Maciej and Blach, Nils and Kubicek, Ales and Gerstenberger, Robert and Gianinazzi, Lukas and Gajda, Joanna and Lehmann, Tomasz and Podstawski, Michal and Niewiadomski, Hubert and Nyczyk, Piotr and others},
  year = {2023},
  journal = {arXiv preprint arXiv:2308.09687},
  doi = {10.48550/arXiv.2308.09687}
}

@misc{cao2024surveylargelanguage,
  title = {Survey on {{Large Language Model-Enhanced Reinforcement Learning}}: {{Concept}}, {{Taxonomy}}, and {{Methods}}},
  author = {Cao, Yuji and Zhao, Huan and Cheng, Yuheng and Shu, Ting and Liu, Guolong and Liang, Gaoqi and Zhao, Junhua and Li, Yun},
  year = {2024},
  publisher = {arXiv},
  doi = {10.48550/arXiv.2404.00282}
}

@inproceedings{chen2023theoremqatheoremdrivenquestion,
    title = "{T}heorem{QA}: A Theorem-driven Question Answering Dataset",
    author = "Chen, Wenhu  and
      Yin, Ming  and
      Ku, Max  and
      Lu, Pan  and
      Wan, Yixin  and
      Ma, Xueguang  and
      Xu, Jianyu  and
      Wang, Xinyi  and
      Xia, Tony",
    editor = "Bouamor, Houda  and
      Pino, Juan  and
      Bali, Kalika",
    booktitle = "Proceedings of the 2023 Conference on Empirical Methods in Natural Language Processing",
    month = dec,
    year = "2023",
    address = "Singapore",
    publisher = "Association for Computational Linguistics",
    url = "https://aclanthology.org/2023.emnlp-main.489/",
    doi = "10.18653/v1/2023.emnlp-main.489",
    pages = "7889--7901",
}

@misc{chenghaoyang2024inferencetimecompute,
  title = {Inference {{Time Compute}}},
  author = {{Chenghao Yang}},
  year = {2024},
  url = {https://www.youtube.com/watch?v=_Bw5o55SRL8}
}

@article{mab_19,
  title     = {Introduction to Multi-Armed Bandits},
  author    = {Aleksandrs Slivkins},
  journal   = {Found. Trends Mach. Learn.},
  year      = {2019},
  doi       = {10.1561/2200000068},
  bibSource = {Semantic Scholar https://www.semanticscholar.org/paper/4c7730d6227f8b90735ba4de7864551cb8928d92}
}

@inproceedings{
didolkar2024metacognitivecapabilitiesllms,
title={Metacognitive Capabilities of {LLM}s: An Exploration in Mathematical Problem Solving},
author={Aniket Rajiv Didolkar and Anirudh Goyal and Nan Rosemary Ke and Siyuan Guo and Michal Valko and Timothy P Lillicrap and Danilo Jimenez Rezende and Yoshua Bengio and Michael Curtis Mozer and Sanjeev Arora},
booktitle={AI for Math Workshop @ ICML 2024},
year={2024},
url={https://openreview.net/forum?id=0MsI3bSmmD}
}

@inproceedings{gandhi2024streamsearchsos,
  title = {Stream of {{Search}} ({{SoS}}): {{Learning}} to {{Search}} in {{Language}}},
  booktitle = {First {{Conference}} on {{Language Modeling}}},
  author = {Gandhi, Kanishk and Lee, Denise H. J. and Grand, Gabriel and Liu, Muxin and Cheng, Winson and Sharma, Archit and Goodman, Noah},
  year = {2024},
  url = {https://openreview.net/forum?id=2cop2jmQVL#discussion},
  langid = {english}
}

@article{gao2024metareasoninglarge,
  title   = {Meta Reasoning for Large Language Models},
  author  = {Peizhong Gao and Ao Xie and Shaoguang Mao and Wenshan Wu and Yan Xia and Haipeng Mi and Furu Wei},
  year    = {2024},
  journal = {arXiv preprint arXiv: 2406.11698}
}

@inproceedings{havrilla2024glorewhenwhere,
  author={Alexander Havrilla and Sharath Chandra Raparthy and Christoforos Nalmpantis and Jane Dwivedi-Yu and Maksym Zhuravinskyi and Eric Hambro and Roberta Raileanu},
  title={GLoRe: When, Where, and How to Improve LLM Reasoning via Global and Local Refinements},
  year={2024},
  cdate={1704067200000},
  url={https://openreview.net/forum?id=LH6R06NxdB},
  booktitle={ICML}
}

@article{lee2025evolvingdeeperllm,
  title   = {Evolving Deeper LLM Thinking},
  author  = {Kuang-Huei Lee and Ian Fischer and Yueh-Hua Wu and Dave Marwood and Shumeet Baluja and Dale Schuurmans and Xinyun Chen},
  year    = {2025},
  journal = {arXiv preprint arXiv: 2501.09891}
}

@inproceedings{
lei2024macmutilizingmultiagent,
title={{MACM}: Utilizing a Multi-Agent System for Condition Mining in Solving Complex Mathematical Problems},
author={Bin Lei and Yi Zhang and Shan Zuo and Ali Payani and Caiwen Ding},
booktitle={The Thirty-eighth Annual Conference on Neural Information Processing Systems},
year={2024},
url={https://openreview.net/forum?id=VR2RdSxtzs}
}

@inproceedings{li2012contextualbanditapproachpersonalized,
author = {Li, Lihong and Chu, Wei and Langford, John and Schapire, Robert E.},
title = {A contextual-bandit approach to personalized news article recommendation},
year = {2010},
isbn = {9781605587998},
publisher = {Association for Computing Machinery},
address = {New York, NY, USA},
url = {https://doi.org/10.1145/1772690.1772758},
doi = {10.1145/1772690.1772758},
booktitle = {Proceedings of the 19th International Conference on World Wide Web},
pages = {661–670},
numpages = {10},
location = {Raleigh, North Carolina, USA},
series = {WWW '10}
}

@article{li2025searcho1agenticsearchenhanced,
  title   = {Search-o1: Agentic Search-Enhanced Large Reasoning Models},
  author  = {Xiaoxi Li and Guanting Dong and Jiajie Jin and Yuyao Zhang and Yujia Zhou and Yutao Zhu and Peitian Zhang and Zhicheng Dou},
  year    = {2025},
  journal = {arXiv preprint arXiv: 2501.05366}
}

@article{lightman2023letsverifystep,
  title     = {Let's Verify Step by Step},
  author    = {Hunter Lightman and Vineet Kosaraju and Yura Burda and Harrison Edwards and Bowen Baker and Teddy Lee and Jan Leike and John Schulman and I. Sutskever and K. Cobbe},
  journal   = {International Conference on Learning Representations},
  year      = {2023},
  doi       = {10.48550/arXiv.2305.20050},
  bibSource = {Semantic Scholar https://www.semanticscholar.org/paper/be8db99310602d66bba64bcf41a572c45816fbfc}
}

@inproceedings{ling2023deductiveverificationchainofthought,
author = {Ling, Zhan and Fang, Yunhao and Li, Xuanlin and Huang, Zhiao and Lee, Mingu and Memisevic, Roland and Su, Hao},
title = {Deductive verification of chain-of-thought reasoning},
year = {2023},
publisher = {Curran Associates Inc.},
address = {Red Hook, NY, USA},
booktitle = {Proceedings of the 37th International Conference on Neural Information Processing Systems},
articleno = {1580},
numpages = {27},
location = {New Orleans, LA, USA},
series = {NIPS '23}
}

@misc{
manvi2024adaptiveinferencetimecompute,
title={Adaptive Inference-Time Compute: {LLM}s Can Predict if They Can Do Better, Even Mid-Generation},
author={Rohin Manvi and Anikait Singh and Stefano Ermon},
year={2025},
url={https://openreview.net/forum?id=7tOc6h8bea}
}

@misc{patel2024aimeaisystem,
  title  = {{AIME}: {AI} System Optimization via Multiple {LLM} Evaluators},
  author = {Bhrij Patel and Souradip Chakraborty and Wesley A. Suttle and Mengdi Wang and Amrit Singh Bedi and Dinesh Manocha},
  year   = {2024},
  url    = {https://openreview.net/forum?id=Z6kVjQAPNq}
}

@inproceedings{
rein2023gpqagraduatelevelgoogleproof,
title={{GPQA}: A Graduate-Level Google-Proof Q\&A Benchmark},
author={David Rein and Betty Li Hou and Asa Cooper Stickland and Jackson Petty and Richard Yuanzhe Pang and Julien Dirani and Julian Michael and Samuel R. Bowman},
booktitle={First Conference on Language Modeling},
year={2024},
url={https://openreview.net/forum?id=Ti67584b98}
}

@inproceedings{
shinn2024reflexion,
title={Reflexion: language agents with verbal reinforcement learning},
author={Noah Shinn and Federico Cassano and Ashwin Gopinath and Karthik R Narasimhan and Shunyu Yao},
booktitle={Thirty-seventh Conference on Neural Information Processing Systems},
year={2023},
url={https://openreview.net/forum?id=vAElhFcKW6}
}

@article{snell2024scalingllmtesttime,
  title   = {Scaling LLM Test-Time Compute Optimally can be More Effective than Scaling Model Parameters},
  author  = {Charlie Snell and Jaehoon Lee and Kelvin Xu and Aviral Kumar},
  year    = {2024},
  journal = {arXiv preprint arXiv: 2408.03314}
}

@inproceedings{sui2024canknowledgegraphs,
  title     = {Can Knowledge Graphs Make Large Language Models More Trustworthy? An Empirical Study Over Open-ended Question Answering},
  author    = {Sui, Yuan and Yufei, He and Zifeng, Ding and Bryan Hooi},
  booktitle = {The 63rd Annual Meeting of the Association for Computational Linguistics},
  year      = {2025},
  url       = {https://openreview.net/forum?id=8gQ5l9DmyY}
}

@inproceedings{wang2022self,
  title = {Self-Consistency Improves Chain of Thought Reasoning in Language Models},
  booktitle = {The Eleventh International Conference on Learning Representations},
  author = {Wang, Xuezhi and Wei, Jason and Schuurmans, Dale and Le, Quoc V and Chi, Ed H and Narang, Sharan and Chowdhery, Aakanksha and Zhou, Denny},
  year = {2022}
}

@article{wang2024scibenchevaluatingcollegelevel,
  title     = {SciBench: Evaluating College-Level Scientific Problem-Solving Abilities of Large Language Models},
  author    = {Xiaoxuan Wang and Ziniu Hu and Pan Lu and Yanqiao Zhu and Jieyu Zhang and Satyen Subramaniam and Arjun R. Loomba and Shichang Zhang and Yizhou Sun and Wei Wang},
  journal   = {International Conference on Machine Learning},
  year      = {2023},
  doi       = {10.48550/arXiv.2307.10635},
  bibSource = {Semantic Scholar https://www.semanticscholar.org/paper/4993258852711c4e3d0011325ac3db680eae84f4}
}

@article{wei2022chain,
  title = {Chain-of-Thought Prompting Elicits Reasoning in Large Language Models},
  author = {Wei, Jason and Wang, Xuezhi and Schuurmans, Dale and Bosma, Maarten and Xia, Fei and Chi, Ed and Le, Quoc V and Zhou, Denny and others},
  year = {2022},
  journal = {Advances in Neural Information Processing Systems},
  volume = {35},
  pages = {24824--24837}
}

@inproceedings{
weng2023largelanguagemodels,
title={Large Language Models are Better Reasoners with Self-Verification},
author={Yixuan Weng and Minjun Zhu and Fei Xia and Bin Li and Shizhu He and Shengping Liu and Bin Sun and Kang Liu and Jun Zhao},
booktitle={The 2023 Conference on Empirical Methods in Natural Language Processing},
year={2023},
url={https://openreview.net/forum?id=s4xIeYimGQ}
}

@inproceedings{
yao2023treethoughtsdeliberate,
title={Tree of Thoughts: Deliberate Problem Solving with Large Language Models},
author={Shunyu Yao and Dian Yu and Jeffrey Zhao and Izhak Shafran and Thomas L. Griffiths and Yuan Cao and Karthik R Narasimhan},
booktitle={Thirty-seventh Conference on Neural Information Processing Systems},
year={2023},
url={https://openreview.net/forum?id=5Xc1ecxO1h}
}

@inproceedings{
yoran2024answeringquestionsmetareasoning,
title={Answering Questions by Meta-Reasoning over Multiple Chains of Thought},
author={Ori Yoran and Tomer Wolfson and Ben Bogin and Uri Katz and Daniel Deutch and Jonathan Berant},
booktitle={The 2023 Conference on Empirical Methods in Natural Language Processing},
year={2023},
url={https://openreview.net/forum?id=ebSOK1nV2r}
}

@misc{guo2025evopromptconnectingllmsevolutionary,
      title={EvoPrompt: Connecting LLMs with Evolutionary Algorithms Yields Powerful Prompt Optimizers}, 
      author={Qingyan Guo and Rui Wang and Junliang Guo and Bei Li and Kaitao Song and Xu Tan and Guoqing Liu and Jiang Bian and Yujiu Yang},
      year={2025},
      eprint={2309.08532},
      archivePrefix={arXiv},
      primaryClass={cs.CL},
      url={https://arxiv.org/abs/2309.08532}, 
}

@misc{wu2025exampleshighlevelautomatedreasoning,
      title={Beyond Examples: High-level Automated Reasoning Paradigm in In-Context Learning via MCTS}, 
      author={Jinyang Wu and Mingkuan Feng and Shuai Zhang and Feihu Che and Zengqi Wen and Chonghua Liao and Jianhua Tao},
      year={2025},
      eprint={2411.18478},
      archivePrefix={arXiv},
      primaryClass={cs.CL},
      url={https://arxiv.org/abs/2411.18478}, 
}

@article{DBLP:journals/corr/abs-2411-15594,
  author       = {Jiawei Gu and
                  Xuhui Jiang and
                  Zhichao Shi and
                  Hexiang Tan and
                  Xuehao Zhai and
                  Chengjin Xu and
                  Wei Li and
                  Yinghan Shen and
                  Shengjie Ma and
                  Honghao Liu and
                  Yuanzhuo Wang and
                  Jian Guo},
  title        = {A Survey on LLM-as-a-Judge},
  journal      = {CoRR},
  volume       = {abs/2411.15594},
  year         = {2024},
  url          = {https://doi.org/10.48550/arXiv.2411.15594},
  doi          = {10.48550/ARXIV.2411.15594},
  eprinttype    = {arXiv},
  eprint       = {2411.15594},
  bibsource    = {dblp computer science bibliography, https://dblp.org}
}

@article{DBLP:journals/corr/abs-2412-05579,
  author       = {Haitao Li and
                  Qian Dong and
                  Junjie Chen and
                  Huixue Su and
                  Yujia Zhou and
                  Qingyao Ai and
                  Ziyi Ye and
                  Yiqun Liu},
  title        = {LLMs-as-Judges: {A} Comprehensive Survey on LLM-based Evaluation Methods},
  journal      = {CoRR},
  volume       = {abs/2412.05579},
  year         = {2024},
  url          = {https://doi.org/10.48550/arXiv.2412.05579},
  doi          = {10.48550/ARXIV.2412.05579},
  eprinttype    = {arXiv},
  eprint       = {2412.05579},
  bibsource    = {dblp computer science bibliography, https://dblp.org}
}

\appendix
\clearpage

\section{Related Works}
\label{sec:related_works}

We first review the landscape of complex reasoning in LLMs, focusing on the transition from static Chain-of-Thought (CoT) to dynamic inference-time search. We then discuss the dual-process systems from cognitive science to frame the design our approach \ours{}.

\paragraph{\textbf{Complex Reasoning \& Inference-Time Compute.}} Standard Chain-of-Thought (CoT)~\cite{wei2022chain} reasoning has been the cornerstone of LLM problem-solving~\cite{lee2025evolvingdeeperllm}. Recent advancements, such as OpenAI's o1/o3 and Deepseek's r1, demonstrate that scaling test-time computation to allow the model to "think" longer, can yield state-of-the-art performance~\cite{manvi2024adaptiveinferencetimecompute,li2025searcho1agenticsearchenhanced}. 
However, CoT relies heavily on a strict step-by-step sequence. Consequently, a single error in an early step often cascades, causing the entire reasoning chain to fail~\cite{snell2024scalingllmtesttime,zhang2025survey}. Furthermore, when models face difficult tasks, they often get stuck in repetitive loops without realizing they are making no progress~\cite{sui2024canknowledgegraphs}. \ours{} addresses these issues by introducing a meta-controller that actively monitors progress and intervenes when the reasoning process stalls or goes off track.

\paragraph{\textbf{Self-Correction and Verification.}} To address the issues of error propagation, recent works have introduced mechanisms for self-correction and backtracking~\cite{yao2023treethoughtsdeliberate,besta2023graphthoughtssolving,gandhi2024streamsearchsos}. Methods like Reflexion~\cite{shinn2024reflexion} and self-verification frameworks~\cite{weng2023largelanguagemodels,ling2023deductiveverificationchainofthought} ask the model to review and revise its own output. While effective, these approaches often rely on fixed heuristics (e.g., "always verify after step $N$") or simple scalar scores that only indicate if a step is "good" or "bad"~\cite{lightman2023letsverifystep,zhang2024learning}. \ours{} improves upon this by using a learned policy using multi-armed bandits to make specific decisions. Instead of just giving a score, our meta-reasoner provides actionable instructions such as "backtrack to the previous step" or "restart with a new method", allowing for more flexible and intelligent error correction than standard verification loops.


\paragraph{\textbf{Meta-Reasoner vs. Tree Search.}} Search algorithms like Tree-of-Thoughts (ToT)~\cite{yao2023treethoughtsdeliberate} and MCTS~\cite{wu2025exampleshighlevelautomatedreasoning} treat reasoning as a search problem, trying to find the best possible next step at every point. However, this approach is computationally expensive because it requires managing a vast number of potential steps. In contrast, \ours{} operates at a higher level: it functions as a \textit{strategy} controller rather than a \textit{step} controller. Instead of micro-managing every word the model generates, our framework periodically evaluates the overall path and selects high-level strategies via Contextual Multi-Armed Bandits. This allows \ours{} to guide the general direction of the solution without the high computational cost of expanding a full search tree.

\paragraph{\textbf{Meta-Cognition \& Dual-Process Systems.}}
From a cognitive science perspective, meta-cognition involves higher-order processes that allow individuals to monitor, evaluate, and adjust their cognitive strategies~\cite{gao2024metareasoninglarge,yoran2024answeringquestionsmetareasoning} during thinking process. This reflective thinking—often characterized as System 2 in dual-process theories~\cite{havrilla2024glorewhenwhere}—is vital for tasks requiring careful deliberation and error correction~\cite{didolkar2024metacognitivecapabilitiesllms}. Drawing on these insights, our \ours{} framework can be viewed as analogous to dual-process systems: the LLM generates CoT steps akin to System 1, while the Meta-Reasoner provides high-level strategic oversight, analogous to System 2, guiding or redirecting reasoning as needed. This separation of responsibilities balances efficiency with robust problem-solving, allowing the LLM to handle routine inferences and the Meta-Reasoner to intervene for strategic adjustments.

\section{Experiment Details}
\label{sec:training_details}

\paragraph{\textbf{Datasets.}} 
We mainly focus on tasks that demand complex reasoning and often involve lengthy thinking processes for the correct solutions. These includes (1) 24-point game~\cite{yao2023treethoughtsdeliberate}; (2) college-level scientific problem from SciBench~\cite{wang2024scibenchevaluatingcollegelevel}; (3) math questions based on theorems from TheoremQA~\cite{chen2023theoremqatheoremdrivenquestion} and (4) math questions from Math-500~\cite{lightman2023letsverifystep} and AIME-2024~\cite{guan2025rstar0math0,aimo2024validationaime}. 
For SciBench, we focus only on the math-related subsets (i.e., diff, stat, and calc). Detailed explanations for each subset can be found in \citet{wang2024scibenchevaluatingcollegelevel}. For TheormQA, we only consider the math subset that involves logical reasoning.

\begin{table}[th]
\centering
\resizebox{\linewidth}{!}{
\begin{tabular}{llcc}
\toprule
\textbf{Model} & \textbf{Method} & \textbf{Token Usage} & \textbf{Inference Time (s)} \\
\midrule
qwen-3-8B & \textbf{Meta-Reasoner} & 1728.9 $\pm$ 42.3 & 31.70 $\pm$ 1.24 \\
qwen-3-8B & MACM & 2266.78 $\pm$ 58.1 & 41.35 $\pm$ 1.87 \\
qwen-3-8B & ToT (b=5) & 2535.72 $\pm$ 67.4 & 46.17 $\pm$ 2.15 \\
qwen-3-8B & Best of N & 2497.3 $\pm$ 63.2 & 45.48 $\pm$ 2.03 \\
qwen-3-8B & Zero-shot & 153.68 $\pm$ 4.2 & 3.47 $\pm$ 0.18 \\
\midrule
o1-preview & Zero-shot & 3534.64 $\pm$ 128.5 & 44.99 $\pm$ 2.67 \\
o1-mini & Zero-shot & 2766.24 $\pm$ 95.3 & 17.05 $\pm$ 1.12 \\
\midrule
meta-llama-3.1-8B & \textbf{Meta-Reasoner} & 1728.9 $\pm$ 38.7 & 8.93 $\pm$ 0.41 \\
meta-llama-3.1-8B & MACM & 2266.78 $\pm$ 54.2 & 11.62 $\pm$ 0.53 \\
meta-llama-3.1-8B & ToT (b=5) & 2535.72 $\pm$ 61.8 & 12.97 $\pm$ 0.62 \\
meta-llama-3.1-8B & Best of N & 2497.3 $\pm$ 59.4 & 12.78 $\pm$ 0.58 \\
meta-llama-3.1-8B & Zero-shot & 153.68 $\pm$ 3.9 & 1.06 $\pm$ 0.05 \\
\midrule
gpt-4o-mini & \textbf{Meta-Reasoner} & 1728.9 $\pm$ 45.1 & 18.60 $\pm$ 0.89 \\
gpt-4o-mini & MACM & 2266.78 $\pm$ 59.7 & 24.30 $\pm$ 1.15 \\
gpt-4o-mini & ToT (b=5) & 2535.72 $\pm$ 68.3 & 27.15 $\pm$ 1.34 \\
gpt-4o-mini & Best of N & 2497.3 $\pm$ 65.8 & 26.74 $\pm$ 1.28 \\
gpt-4o-mini & Zero-shot & 153.68 $\pm$ 4.5 & 1.92 $\pm$ 0.09 \\
\midrule
ds-r1-distill-qwen-7B & \textbf{Meta-Reasoner} & 1728.9 $\pm$ 41.5 & 17.20 $\pm$ 0.82 \\
ds-r1-distill-qwen-7B & MACM & 2266.78 $\pm$ 56.3 & 21.26 $\pm$ 1.03 \\
ds-r1-distill-qwen-7B & ToT (b=5) & 2535.72 $\pm$ 65.1 & 23.29 $\pm$ 1.18 \\
ds-r1-distill-qwen-7B & Best of N & 2497.3 $\pm$ 62.7 & 23.00 $\pm$ 1.12 \\
ds-r1-distill-qwen-7B & Zero-shot & 153.68 $\pm$ 4.1 & 5.33 $\pm$ 0.28 \\
\midrule
ds-r1-distill-llama-8B & \textbf{Meta-Reasoner} & 1728.9 $\pm$ 43.8 & 31.49 $\pm$ 1.52 \\
ds-r1-distill-llama-8B & MACM & 2266.78 $\pm$ 57.9 & 40.98 $\pm$ 1.96 \\
ds-r1-distill-llama-8B & ToT (b=5) & 2535.72 $\pm$ 66.7 & 45.72 $\pm$ 2.21 \\
ds-r1-distill-llama-8B & Best of N & 2497.3 $\pm$ 64.3 & 45.04 $\pm$ 2.15 \\
ds-r1-distill-llama-8B & Zero-shot & 153.68 $\pm$ 4.3 & 3.70 $\pm$ 0.19 \\
\bottomrule
\end{tabular}}
\caption{Inference Compute Across Different Methods (Mean $\pm$ StdDev over 3 runs)}
\end{table}

\paragraph{\textbf{Training Details.}}
We collect the training data for each task: (1) for the 24-point game, we random sample 50 queries from 4nums.com specifically excluding problems in ranks 901-1000 which were reserved for testing; (2) for TheoremQA, we randomly sample 30 mathematical reasoning queries from the dataset; (3) for SciBench, we randomly sample 30 queries from differential subsets including diff, stat, and calc from the entire dataset. These samples were used to iteratively update the LinUCB parameters for both fixed ($K$=3 or $K$=5 strategies) and dynamic strategy settings. We configure the training process using deterministic generation ($n$=1, $\text{Top\_k}$=1, temperature=0) with specific $\text{max\_token}$ limits for CoT generation (512), meta-reasoner feedback (256), progress reports (512), and reward model outputs (4). The LinUCB exploration parameter was set to $c$=0.2. Experiments were ran for a maximum of $T$=30 iterations for the 24-point game, scibench, and $T$=100 for TheoremQA, with MAB parameters updated after each iteration based on a reward function that weighted objective completion (40\%), progress quality (30\%), efficiency (15\%), and strategy alignment (15\%). The reward function prompt is detailed in Appendix \refsec{appendix:prompt_list}.

\begin{figure}[ht]
     \centering
    \begin{subfigure}[b]{0.48\linewidth}
         \centering
        \includegraphics[width=1\textwidth,trim={0cm 0.2cm 0cm 0cm}]{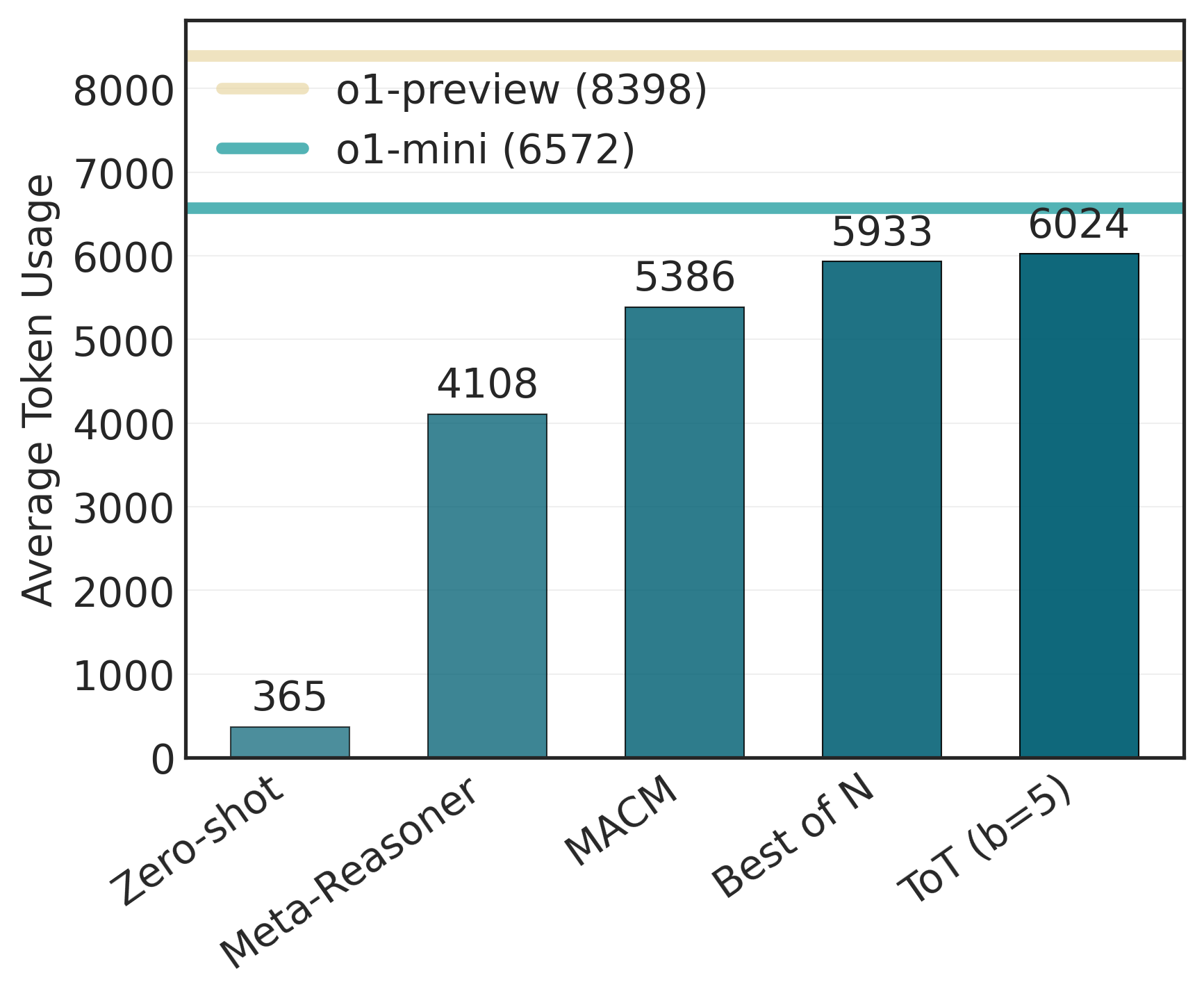}
         \caption{Game of 24 Task.}
     \end{subfigure}
     \hfill
     \hfill
     \hfill
     \hfill 
     \begin{subfigure}[b]{0.48\linewidth}
         \centering
        \includegraphics[width=1\textwidth,trim={0cm 0.2cm 0cm 0cm}]{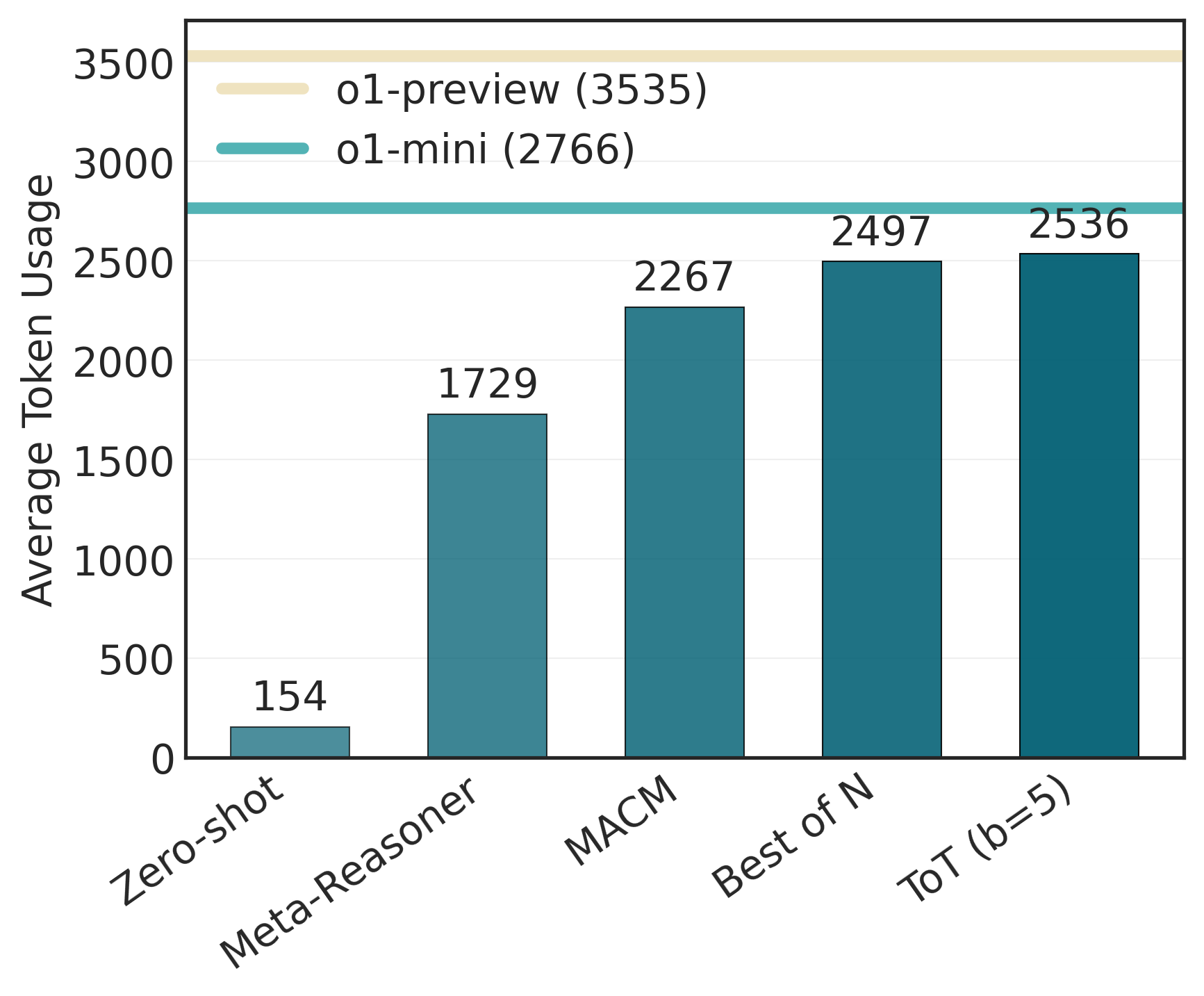}
         \caption{TheoremQA Task}
     \end{subfigure}
     \caption{Token Usage on Game-of-24~\cite{yao2023treethoughtsdeliberate} and TheoremQA~\cite{chen2023theoremqatheoremdrivenquestion} Tasks.}
     \label{fig:token_usage}
\end{figure}

\begin{table}[h]
\centering
\resizebox{\linewidth}{!}{
\begin{tabular}{llccccc}
\toprule
\textbf{Model} & \textbf{Method} & \textbf{Game-of-24} & \textbf{TheoremQA} & \textbf{Diff (\%)} & \textbf{Stat (\%)} & \textbf{Calc (\%)} \\
\midrule
gpt-4o-mini & \textbf{Meta-reasoner} & 89 & 84.13 & 60.32 & 73.64 & 80.23 \\
gpt-4o-mini & HiAR-ICL~\cite{wu2024beyond} & 87 & 83.48 & 57.42 & 70.12 & 77.93 \\
gpt-4o-mini & Evo-Prompt~\cite{guo2023evoprompt0} & 82 & 81.28 & 55.28 & 67.32 & 76.53 \\
\midrule
gemini-exp-1206 & \textbf{Meta-reasoner} & 94 & 86.32 & 65.47 & 79.42 & 82.77 \\
gemini-exp-1206 & HiAR-ICL~\cite{wu2024beyond} & 88 & 84.41 & 57.76 & 75.92 & 80.23 \\
gemini-exp-1206 & Evo-Prompt~\cite{guo2023evoprompt0} & 84 & 80.32 & 57.32 & 70.32 & 78.42 \\
\bottomrule
\end{tabular}}
\caption{Comparison with HiAR-ICL~\cite{wu2025exampleshighlevelautomatedreasoning} and Evo-Prompt~\cite{guo2025evopromptconnectingllmsevolutionary}.}
\end{table}

\begin{table*}[t]
\centering
\small
\resizebox{0.75\linewidth}{!}{
\begin{tabular}{lcccc}
\toprule
\textbf{Parameter Variation} & \textbf{Game-of-24 Acc. (\%)} & \textbf{TheoremQA Acc. (\%)} \\
\midrule
\multicolumn{5}{c}{\textit{Default: $w_1=0.5, w_2=0.5, \alpha=0.1, \beta=0.8$}} \\
Default & 89.0 $\pm$ 1.2 & 84.1 $\pm$ 1.5 \\
\midrule
\multicolumn{5}{c}{\textit{Varying $w_1$ and $w_2$ ($\alpha=0.1, \beta=0.8$)}} \\
$w_1=0.3, w_2=0.7$ (Adherence emphasis) & 87.5 $\pm$ 1.4 & 83.2 $\pm$ 1.6 \\
$w_1=0.7, w_2=0.3$ (Correctness emphasis) & 88.2 $\pm$ 1.3 & 85.3 $\pm$ 1.4  \\
$w_1=0.9, w_2=0.1$ (Heavy correctness) & 87.8 $\pm$ 1.5 & 84.8 $\pm$ 1.7 \\
\midrule
\multicolumn{5}{c}{\textit{Varying $\alpha$ ($w_1=0.5, w_2=0.5, \beta=0.8$)}} \\
$\alpha=0.05$ (Low cost penalty) & 89.4 $\pm$ 1.1 & 84.5 $\pm$ 1.4 \\
$\alpha=0.2$ (Moderate penalty) & 88.6 $\pm$ 1.3 & 83.7 $\pm$ 1.5\\
$\alpha=0.5$ (High penalty) & 84.2 $\pm$ 1.6 & 79.8 $\pm$ 1.8 \\
\midrule
\multicolumn{5}{c}{\textit{Varying $\beta$ ($w_1=0.5, w_2=0.5, \alpha=0.1$)}} \\
$\beta=0.6$ (Balanced trade-off) & 88.1 $\pm$ 1.2 & 83.4 $\pm$ 1.5 \\
$\beta=0.5$ (Efficiency emphasis) & 86.7 $\pm$ 1.4 & 82.0 $\pm$ 1.7  \\
$\beta=0.9$ (Progress emphasis) & 89.3 $\pm$ 1.1 & 84.6 $\pm$ 1.4 \\
\bottomrule
\end{tabular}}
\caption{Sensitivity analysis results on reward function weights.}
\label{tab:sensitivity}
\end{table*}

\paragraph{\textbf{Baselines.}}
We consider several established prompting methods as baselines as follows: (1) Chain-of-thought (CoT)~\cite{wei2022chain}: A prompting technique that encourages models to generate intermediate reasoning steps to enhance problem-solving capabilities. (2) Self-Consistent Chain of Thought (SC-CoT)~\cite{wang2022self}: An extension of CoT that improves reasoning consistency by generating multiple reasoning chains and selecting the most consistent answer. (3) Multi-Chain Reasoning (MCR)~\cite{yoran2024answeringquestionsmetareasoning}: enhances SC-CoT by having another LLM to assess and integrate content among the sampled reasoning chains to generate the final consistent answer.
(4) Tree of Thoughts (ToT)~\cite{yao2023treethoughtsdeliberate}: A method that explores multiple reasoning paths in a tree structure, allowing the model to consider various possibilities before arriving at a conclusion by tree search algorithms.
(5) Reflexion~\cite{shinn2024reflexion}: A framework that enables models to reflect on their reasoning process, iteratively refining their answers based on feedback.
(6) MACM~\cite{lei2024macmutilizingmultiagent}: A multi-agent system to refine the reasoning based on iterative condition mining.
(7) HiAR-ICL~\cite{wu2025exampleshighlevelautomatedreasoning}: A high-level automated reasoning paradigm for in-context learning that constructs abstract thinking patterns using Monte Carlo Tree Search (MCTS) on atomic actions, dynamically matching them to problems via a cognitive complexity framework to reduce reliance on specific examples.
(8) Evo-Prompt~\cite{guo2025evopromptconnectingllmsevolutionary}: A discrete prompt optimization framework that integrates LLMs with evolutionary algorithms, leveraging LLMs to generate coherent prompt variants through operators like crossover and mutation for iterative improvement without gradients.

\paragraph{\textbf{Backbone Models.}}
We consider both LLMs and the recent Large Reasoning Models (LRMs) for our experiments.
For the LLMs, we consider the closed-source models like gpt-4o, gpt-4o-mini (between Nov 2025 to Jan 2025) from OpenAI, and open-sourced models like meta-llama-3.1-8B-instruct from Meta, qwen-3-8B from Alibaba, phi-4 from Microsoft, gemini-experimental-1206 from Google, ds-r1-distill-llama-8B, ds-r1-distill-qwen-7B from Deepseek. For the LRMs, we consider the closed-source models like o1, o1-mini (In case we cannot break down the generation of o1 models through APIs, we cannot properly inject our meta-reasoner with o1-series models; we only provide the IO results for references). For the feature extraction mentioned in \refsec{sec:strategy_generation}, we use text-embedding-3-small from OpenAI as the embedding model. To ensure the reproducibility of the experiments, we set $\mathrm{temperature}=0.7$ and $\mathrm{top\_p}=1.0$ for all models. We use the API service from OpenAI\footnote{\url{https://openai.com/}} and OpenRouter\footnote{\url{https://openrouter.ai/}} for our experiments which host detailed snapshots of the utilized model versions.

\paragraph{\textbf{Judge Model Analysis.}}
\label{appendix:judge_model}
To justify our choice of gemini-2.5-flash as the reward model, we compared it against a reasoning-capable model, gemini-2.5-pro. We randomly sampled 50 cases from Game-of-24 and TheoremQA to test. 
The results in Table~\ref{tab:judge_comparison} show that the lightweight judge is sufficient for evaluating progress reports, offering the best trade-off between efficiency and performance.

\begin{table}[h]
\centering
\small
\resizebox{\linewidth}{!}{
\begin{tabular}{lccc}
\toprule
\textbf{Judge Model} & \textbf{Game-24 Acc} & \textbf{TheoremQA Acc} & \textbf{Cost/1K calls} \\
\midrule
Gemini-2.5-Flash & 88.0\% & 83.5\% & \$0.015 \\
Gemini-2.5-Pro & 88.5\% & 84.1\% & \$0.125 \\
\midrule
\textbf{Difference} & +0.5\% & +0.6\% & +733\% \\
\bottomrule
\end{tabular}}
\caption{Comparison of Judge Models with Reasoning Mode. Using a stronger reasoning model yields negligible accuracy gains ($<1\%$) but significantly increases cost and latency.}
\label{tab:judge_comparison}
\end{table}

\section{Sensitivity Analysis of Reward Function Weights}
\label{appendix:sensitivity_analysis}

To address concerns about the choice of hyperparameters in the reward function (Section~\ref{sec:reward_modeling}), we perform a sensitivity analysis on the main weights. These include $w_1$ and $w_2$ (weights for correctness $C_c$ and adherence $C_a$ in solution progress $S_p$), $\alpha$ (cost coefficient for resource usage $R_u$), and $\beta$ (weight balancing $S_p$ and $R_u$ in the total reward $R$. In the main experiments, we use the following default values: $w_1 = w_2 = 0.5$ (equal emphasis on correctness and adherence), $\alpha = 0.1$ (moderate penalty for extra steps), and $\beta = 0.8$ (strong preference for progress over cost, since accuracy is the primary objective). We select these defaults using a small grid search on a validation subset comprising 10\% of the Game-of-24 dataset. The goal is to favor accurate solutions while still encouraging efficient inference.

We then evaluate the robustness of the meta-reasoner under different hyperparameter configurations. We vary the weights and measure accuracy on two benchmarks: Game-of-24 and TheoremQA. For each configuration, we run the full meta-reasoner pipeline (with GPT-4o-mini as the backbone) using 5 random seeds. We report mean accuracy and standard deviation. To isolate the effect of the reward function, we use the fixed contextual bandit variant and disable dynamic strategy generation. We consider three groups of variants:
\begin{enumerate}
\item Vary $w_1$ and $w_2$ while keeping $\alpha$ and $\beta$ fixed (to test the balance between correctness and adherence).
\item Vary $\alpha$ (to test sensitivity to penalties on computational cost).
\item Vary $\beta$ (to test the trade-off between progress and efficiency in the total reward).
\end{enumerate}

\begin{table}[th]
\centering
\resizebox{\linewidth}{!}{
\begin{tabular}{lcccc}
\toprule
\multirow{2}{*}{\textbf{Method}} & \multicolumn{2}{c}{\textbf{TheoremQA}} & \multicolumn{2}{c}{\textbf{Game-of-24}} \\
\cmidrule(lr){2-3} \cmidrule(lr){4-5}
 & \textbf{Acc (\%)} & \textbf{Time (s)} & \textbf{Acc (\%)} & \textbf{Time (s)} \\
\midrule
Full Method & 84.13 & 18.60 & 89.0 & 43.8 \\
w/o MAB (Direct Selection) & 80.74 & 18.17 & 82.0 & 42.9 \\
\midrule
\textbf{Difference} & \textbf{-3.39} & \textbf{-0.43s (-2.3\%)} & \textbf{-7.0} & \textbf{-0.9s (-2.1\%)} \\
\bottomrule
\end{tabular}}
\caption{Impact of CMAB on computational overhead and performance using \texttt{gpt-4o-mini}. The MAB component introduces negligible latency (<3\%) while significantly boosting accuracy.}
\label{tab:cmab_overhead}
\end{table}

Table~\ref{tab:sensitivity} reports the results. Performance remains stable across a broad range of weight values. Even in extreme settings, accuracy degrades by at most 5-7\% relative to the default configuration. For example, increasing the emphasis on correctness ($w_1 = 0.7$, $w_2 = 0.3$) yields a small accuracy gain on TheoremQA (+1.2\%), which likely reflects that task’s focus on logical precision, while equal weighting works best on Game-of-24. A large cost coefficient (e.g., $\alpha = 0.5$) reduces the number of reasoning steps by 20-25\% but harms accuracy on long-horizon tasks such as TheoremQA, which supports the use of moderate efficiency penalties. A lower value $\beta = 0.5$ shifts the reward towards efficiency, giving similar accuracy with 15-20\% fewer tokens on average; this configuration can be useful in resource-constrained settings. Overall, these findings show that the benefits of the meta-reasoner do not rely on precise hyperparameter tuning. The main gains come from the architecture itself, in particular dynamic strategy selection via CMABs and the progress-report interface. Even with suboptimal weights, our method outperforms strong baselines such as MACM (80\ on Game-of-24) and Reflexion (74.32\% on TheoremQA).

\section{Stability of Dynamic Strategy Generation.}
\label{sec:statbility_of_dynamic_strategy}

Dynamic expansion and refinement of the strategy set (Section~\ref{sec:strategy_generation}) increase adaptability. At the same time, they introduce a risk of instability, since naive strategy generation can be random or ungrounded. We address this risk with three stabilizing mechanisms.

(1) \textit{Initial stable foundation}. We begin with a curated set of verified strategies (e.g., Table~\ref{tab:contextual_bandit_demonstration}). Examples include $g_1 =$ “Pause to clarify and disambiguate reasoning” and $g_2 =$ “Decompose the task into sub-tasks.” This set provides a strong baseline. During inference, the LLM proposes new strategies $g_t$ based on the current progress report $P_t$. The proposal is constrained by the task context and the report content, so the system does not add arbitrary or irrelevant strategies.

(2) \textit{Exploration–exploitation balance}. The CMAB uses an $\epsilon$-greedy policy to balance exploration and exploitation. At meta-reasoning round $t$, the probability of exploring a newly added arm $a_t \in G_t \setminus G_{t-1}$ is
\(\epsilon_t = \frac{1}{t}.\)
Thus, exploration is frequent at early rounds and gradually decays. We filter suboptimal strategies during inference based on their empirical reward $R_t(a_t, P_t)$. On the Game-of-24 task, this dynamic bandit achieves 89\% accuracy, whereas fixed strategy sets yield 65\%–72\% accuracy (Table~\ref{tab:fixed_dynamic}). These results indicate that the policy effectively prioritizes viable strategies.

(3) \textit{Reward-driven feedback}. The reward function
\(R: G_t \times P_t \to [0, 1]\)
(defined in Section~\ref{sec:strategy_generation}) provides online feedback on strategy quality. Strategies that consistently yield low rewards (e.g., $R_t < 0.3$) are rapidly down-weighted and effectively removed from future consideration. Figure~\ref{fig:accumulated_rewards} shows that the cumulative reward $\sum_{i=1}^t R_i$ increases over time as the controller incorporates new, useful strategies. This pattern reflects a stable and improving reasoning process.

These three mechanisms act together. Context-constrained generation anchors new strategies in the current reasoning state. Bandit optimization filters and reorders strategies based on observed rewards. Reward-driven feedback then refines the strategy set over time. Empirical results on Game-of-24 and TheoremQA (Table~\ref{tab:fixed_dynamic}) show that dynamic strategy generation yields 17\%–24\% gains over static strategies, while maintaining stable behavior.

\section{Computational Overhead of CMAB}
\label{appendix:cmab_overhead}

To assess the latency introduced by the meta-reasoning module, we compare the computational cost of the CMAB with the cost of base LLM inference. The CMAB uses the LinUCB algorithm and \texttt{text-embedding-3-small} for feature extraction. Both components run in a few milliseconds per meta-reasoning step. In contrast, the main cost comes from the LLM, which generates hundreds of tokens and often takes several seconds. The additional latency from CMAB is therefore small relative to the total inference time. Table~\ref{tab:cmab_overhead} reports a trade-off analysis with \texttt{gpt-4o-mini}. Adding the CMAB increases total inference time by less than 3\% (about 0.4–0.9 seconds for tasks that take 18–44 seconds). At the same time, it improves accuracy by 3–7 percentage points. These results show that the meta-reasoner using CMAB incurs minimal computational overhead while providing clear performance gains.

\newpage
\onecolumn

\section{Prompt List}
\label{appendix:prompt_list}

\begin{tcolorbox}[
  enhanced,
  breakable,
  colback=ysshallowblue,
  colframe=ysdarkblue,
  title={Prompt for Progress Report (\refsec{sec:progress_report})},
  fonttitle=\bfseries
]
\begin{Verbatim}[
  breaklines=true,      % 启用自动换行
  breaksymbolleft=,     % 隐藏换行符号
  showspaces=false,     % 不显示空格
  fontsize=\small,      % 调整字体大小
  commandchars=\\\{\}   % 避免冲突符号
]
You are an advanced AI summarizer with expertise in extracting and condensing key insights from recent developments. Your goal is to create a concise progress report based on the provided information.

Read the task description and the chain of thoughts generated so far. Please ignore the <examples> section which is only for the demonstration of the task.
<progress>
{SOLUTION}
</progress>

And then complete the following template:

Current Attempts:
[Insert the list of previous attempts here]

Analysis Instructions:
1. Systematically review each attempted step
2. Identify patterns in the current solution attempts
3. Provide observations regarding:
   - Recurring strategies
   - Missed opportunities
   - Potential promising approaches
   - Any mathematical observations about the number combination

Output Format:
- Provide a structured analysis
- Include bullet points for key observations

Constraints:
- Use clear, logical reasoning
- Focus on mathematical problem-solving approaches
- Avoid random guessing
- Make the analysis short and to the point (around 6-7 sentences)
\end{Verbatim}
\end{tcolorbox}

\begin{tcolorbox}[
  enhanced,
  breakable,
  colback=ysshallowblue,
  colframe=ysdarkblue,
  title=Prompt for Meta-Reasoner (Dynamic Bandit Generation (\refsec{method:dynamic_contextual_bandits}),
  fonttitle=\bfseries
]
\begin{Verbatim}[
  breaklines=true,      % 启用自动换行
  breaksymbolleft=,     % 隐藏换行符号
  showspaces=false,     % 不显示空格
  fontsize=\small,      % 调整字体大小
  commandchars=\\\{\}   % 避免冲突符号
]
You are a Meta-reasoner, tasked with analyzing the reasoning process of another agent and providing guidance for its further steps. Your goal is to improve the efficiency and effectiveness of that agent's problem-solving approach.

Review the task description and the summary of the recent reasoning progress below:
{PROGRESS_REPORT}

Provide feedback in the following format:
- Reflection: What is the current strategy of the agent to solve the task? Has the agent made sufficient progress? Are there any mistakes or misconceptions in the intermediate steps? Is the agent taking unnecessary detours or repeating steps?
- Fact Check: Are the agent's statements accurate and relevant to the task? Are there any logical errors or incorrect assumptions?
- Thought: What are the key insights or strategies that the agent should focus on? Are there alternative methods or perspectives that could be beneficial?
- Action: The action to take

Make your response precise and focused without unnecessary details.
\end{Verbatim}
\end{tcolorbox}

\begin{tcolorbox}[
  enhanced,
  breakable,
  colback=ysshallowblue,
  colframe=ysdarkblue,
  title={Prompt for CoT Generation (\refsec{sec:strategy_generation})},
  fonttitle=\bfseries
]
\begin{Verbatim}[
  breaklines=true,      % 启用自动换行
  breaksymbolleft=,     % 隐藏换行符号
  showspaces=false,     % 不显示空格
  fontsize=\small,      % 调整字体大小
  commandchars=\\\{\}   % 避免冲突符号
]
You are an AI assistant tasked with generating steps to solve mathematical problems. Your role is to read a task description, consider the current step (if any), and generate the next logical step towards solving the problem. You will also receive feedback from a Meta-reasoner, which you should take into account when determining your next step.

Here is the task description:
<task_description>
{TASK_DESCRIPTION}
</task_description>

The process will work as follows:
1. You will be given the current step (if any) in the problem-solving process.
2. You will also receive feedback from the Meta-reasoner about the previous step.
3. Your job is to generate the next logical step towards solving the problem, taking into account the task description, the current step, and the Meta-reasoner's feedback.

To generate the next step:
1. Carefully analyze the task description, the current step (if any), and the Meta-reasoner's feedback.
2. If the Meta-reasoner suggests backtracking, consider how to modify or correct the previous step.
3. If the Meta-reasoner suggests continuing, think about the logical progression from the current step.
4. If the Meta-reasoner suggests changing strategy, brainstorm alternative approaches to the problem.
5. Formulate a clear, concise next step that moves towards solving the problem.

Your response should be a single, well-thought-out step that progresses the problem-solving process. Do not solve the entire problem at once; focus on generating just the next logical step.

Please provide your next step within <next_step> tags. Before giving your next step, explain your reasoning within <reasoning> tags. Explicitly state whether the problem is solved or not before providing the next step or final answer.
If you believe there has been enough progress to solve the problem completely, generate the final answer in the form of \\boxed{{answer}} at the end of your response. The answer should be a numerical value.

Your response should follow this structure:

<reasoning>
[Explain your thought process here, considering the task description, current step, and Meta-reasoner feedback (make sure to address any issues raised by the Meta-reasoner).
The reasoning should be clear, logical, and directly related to the problem-solving process.]
</reasoning>

<next_step>
[Provide the next logical step here]
</next_step>

[State whether the problem is solved or not]

[If the problem is solved] Return only the Final answer: \\boxed{{numerical_value}}

Remember to focus on generating just the next logical step, not solving the entire problem at once (unless you've reached the final solution). Your explanation and step should be clear, concise, and directly contribute to solving the mathematical problem at hand.

Here is the current step (if this is the first step, this will be empty):
<current_step>
{CURRENT_STEP}
</current_step>

And here is the feedback from the Meta-reasoner (if this is the first step, this will be empty):
<meta_reasoner_feedback>
{META_REASONER_FEEDBACK}
</meta_reasoner_feedback>

\end{Verbatim}
\end{tcolorbox}

\begin{tcolorbox}[
  enhanced,
  breakable,
  colback=ysshallowred,
  colframe=ysdarkred,
  title={Prompt for Progress Evaluation (\refsec{sec:progress_report})},
  fonttitle=\bfseries,
]
\begin{Verbatim}[
  breaklines=true,      % 启用自动换行
  breaksymbolleft=,     % 隐藏换行符号
  showspaces=false,     % 不显示空格
  fontsize=\small,      % 调整字体大小
  commandchars=\\\{\}   % 避免冲突符号
]
You are an impartial evaluator tasked with assessing the progress of a reasoning process toward solving a given task objective. Your evaluation must be based strictly on the provided reward function components. Do not favor any particular output or introduce bias—evaluate objectively.

# Inputs:

Task Objective (G_t): [INSERT TASK OBJECTIVE HERE]  // e.g., the original user query or problem statement

Current Progress (P_t): [INSERT CURRENT REASONING/PROGRESS HERE]  // e.g., the model's accumulated reasoning steps, partial solution, or plan up to this point

Number of Reasoning Steps (N_s): [INSERT NUMBER OF STEPS HERE]  // e.g., the count of iterative reasoning steps taken so far

Weights and Coefficients:
  - w1 (weight for correctness): 0.5
  - w2 (weight for adherence): 0.5
  - alpha (cost coefficient for resource usage): 0.1
  - beta (trade-off balance): 0.8

# Evaluation Criteria:
1. Correctness (C_c): Score on a scale of 0.0 to 1.0 how accurate and logically sound the current progress is toward fully solving the task objective. Consider factual accuracy, logical consistency, and advancement toward a complete solution. 0.0 means no progress or entirely incorrect; 1.0 means perfectly correct and on track for completion.
2. Adherence (C_a): Score on a scale of 0.0 to 1.0 how well the current progress follows the task objective's constraints, requirements, and guidelines (e.g., format, scope, ethical considerations). 0.0 means complete disregard; 1.0 means full compliance.
3. Solution Progress (S_p): Compute as S_p = (w1 * C_c) + (w2 * C_a).
4. Resource Usage (R_u): Compute as R_u = -alpha * N_s. This penalizes excessive steps for efficiency.
5. Total Reward (R): Compute as R = (beta * S_p) + ((1 - beta) * R_u).

# Output Format:
Respond only in the following strict JSON structure. Do not include any additional text, explanations, or commentary outside this JSON.

\{
  "C_c": <float, your score for correctness>,
  "C_a": <float, your score for adherence>,
  "S_p": <float, computed solution progress>,
  "R_u": <float, computed resource usage>,
  "R": <float, total reward>,
  "brief_rationale": "<A concise 1-2 sentence explanation for C_c and C_a scores only.>"
\}

\end{Verbatim}
\label{prompt-progress-evaluation}
\end{tcolorbox}

\end{document}